\documentclass[twoside]{article}

\usepackage[round,authoryear]{natbib}

\usepackage[accepted]{style/aistats2021}

\usepackage{booktabs}       
\usepackage{nicefrac}       
\usepackage{microtype}      

\usepackage{enumitem}

\usepackage{titletoc}

\usepackage[unicode=true,bookmarks=true,bookmarksnumbered=true,bookmarksopen=true,bookmarksopenlevel=1,breaklinks=false,pdfborder={0 0 0},pdfborderstyle={},backref=section,colorlinks=true]{hyperref}

\usepackage{float}

\usepackage{graphicx}
\usepackage{amsmath,amssymb,mathtools,bm,etoolbox,amsthm}
\usepackage{caption}
\usepackage{subcaption}
\usepackage{xcolor}
\usepackage{wrapfig}
\usepackage[title]{appendix}
\usepackage{bm} 

\definecolor{citecol}{rgb}{0,0.5,0}
\definecolor{linkcol}{rgb}{0.8,0,0}
\definecolor{blcolor}{rgb}{0,0,0.65}
\hypersetup{
    colorlinks,
    linkcolor={blcolor},
    citecolor={blcolor},
    urlcolor={blcolor}
}

\usepackage{tikz}
\usetikzlibrary{fit,positioning}

\newcommand{\remove}[1]{}

\newcommand{\bfv}[1]{{\bm{#1}}}

\newcommand{\deriv}[2]{\frac{\textup{d}#1\hfill}{\textup{d}#2\hfill}}

\newcommand{\pderiv}[3]{\left.\frac{\partial#1\hfill}{\partial#2\hfill}\right\vert_{#3}}
\newcommand{\pderivsq}[4]{\left.\frac{\partial^2#1}{\partial#2\partial#3\hfill}\right\vert_{#4}}
\newcommand{\pderivw}[2]{\frac{\partial#1\hfill}{\partial#2\hfill}}
\newcommand{\pderivwsq}[3]{\frac{\partial^2#1\hfill}{\partial#2\partial#3\hfill}}
\newcommand{\expect}[1]{\mathbb{E}\left[#1\right]}
\newcommand{\expectw}[2]{\mathbb{E}_{#1}\left[#2\right]}

\newcommand{\variancew}[2]{\mathbb{V}_{#1}\left[#2\right]}

\newcommand{\p}[1]{p\left(#1\right)}
\newcommand{\pind}[2]{p_{#1}\left(#2\right)}
\newcommand{\dpx}{\deriv{p}{\bfv{x}}}
\newtheorem{theorem}{Theorem}
\usepackage{xr}
\usepackage{placeins}

\begin{document}

\runningtitle{A unified view of likelihood ratio and reparameterization gradients}

\twocolumn[
\aistatstitle{A unified view of likelihood ratio and reparameterization gradients}
\aistatsauthor{Paavo Parmas \And Masashi Sugiyama}
\aistatsaddress{Kyoto University\footnotemark \And  RIKEN and The University of Tokyo}]
%

\footnotetext{Work mostly performed while affiliated with the Okinawa Institute of Science and Technology and partially performed while interning at RIKEN.}

\begin{abstract}
  Reparameterization (RP) and likelihood ratio (LR) gradient
  estimators are used to estimate gradients of expectations throughout
  machine learning and reinforcement learning; however, they are usually
  explained as simple mathematical tricks, with no insight into their
  nature. We use a first principles approach to explain that LR and RP
  are alternative methods of keeping track of the movement of
  probability mass, and the two are connected via the divergence
  theorem. Moreover, we show that the space of all possible estimators
  combining LR and RP can be completely parameterized by a flow field
  $\bfv{u}(\bfv{x})$ and an importance sampling distribution
  $q(\bfv{x})$. We prove that there cannot exist a single-sample
  estimator of this type outside our characterized space, thus,
  clarifying where we should be searching for better Monte Carlo
  gradient estimators.


\end{abstract}

\section{INTRODUCTION}
\label{intro}
Both likelihood ratio (LR) gradients
\citep{glynn1990likelihood,williams1992reinforce} and
reparameterization (RP) gradients
\citep{rezende2014stochasticBP,kingma2013autoencoder} give
unbiased estimates of the gradient of an expectation w.r.t.\
the parameters of the distribution:
$\deriv{}{\theta}\expectw{\p{x;\theta}}{\phi(x)}$. This gradient
estimation problem
is fundamental in machine learning \citep{mcgradrev}, where the
gradients are used for optimization. LR is the basis
of many reinforcement learning (RL)
\citep{sutton1998reinforcement,schulman2015trpo,schulman2017ppo,sutton2000policy,peters2008polgrad}
and evolutionary algorithms
\citep{wierstra2008natural,salimans2017oaies,ha2018worldmodels,conti2018improving}.
In RL, $\phi(x)$ represents the sum of rewards, and $\theta$ are the
policy parameters.  RP, on the other hand, is the backbone of
stochastic variational inference \citep{hoffman2013stochastic}, where
$\phi(x)$ is the evidence lower bound, and $\theta$ are the
variational parameters. For example, RP is used in autoencoders
\citep{kingma2013autoencoder}.  There is a vast body of research on
both estimators (App.~\ref{additionalbackground}), and there is no
clear winner among RP and LR---both have advantages and disadvantages.
\remove{ LR uses samples of the function values
  $\phi(x)$ to estimate the gradient, and it is even applicable when
  $x$ is discrete. On the other hand, RP uses samples of the gradient
  of the function $\deriv{\phi(x)}{x}$, and is typically said to be
  more accurate and scale better with the sampling dimension
  \citep{rezende2014stochasticBP}; however, {\bf there is no guarantee
    that RP outperforms LR}. In particular, for multimodal $\phi(x)$
  \citep{gal2016uncertainty} or chaotic systems \citep{pipps}, LR can be
  arbitrarily better than RP (e.g., the latter showed that LR can be
  $10^6$ more accurate in practice).  }
\remove{Despite the ubiquitous use of
LR and RP in machine learning, their relationship is not fully
understood.}

LR uses samples of the value of the function $\phi(x)$ to
estimate the gradient, and is usually derived as
\begin{equation}
  \begin{aligned}
    \label{LRderiv}
  \deriv{}{\theta}\expectw{\p{x;\theta}}{\phi(x)} &=
\int\deriv{\p{x;\theta}}{\theta}\phi(x)\textup{d}x \\
&= \int\p{x;\theta}\frac{1}{\p{x;\theta}}\deriv{\p{x;\theta}}{\theta}
\phi(x)\textup{d}x \\ &=
\int\p{x;\theta}\deriv{\log\p{x;\theta}}{\theta}\phi(x)\textup{d}x \\
&=
\expectw{\p{x;\theta}}{\deriv{\log\p{x;\theta}}{\theta}\phi(x)}.
\end{aligned}
\end{equation}
On the other hand, RP uses samples of the gradient of the function
$\deriv{\phi(x)}{x}$, and it is derived by defining a mapping
$g(\epsilon;\theta) = x$, where $\epsilon$ is sampled from a fixed simple
distribution, $p(\epsilon)$, independent of $\theta$, but $x$ ends up being sampled from the desired
distribution. For example, if $x$ is Gaussian,
$x~\sim~\mathcal{N}(\mu,\sigma)$, then the required mapping is
$g(\epsilon;\theta) = \mu +\sigma\epsilon$, where
$\epsilon~\sim~\mathcal{N}(0,1)$, and the RP gradient is derived as
\begin{equation}
  \begin{aligned}
    \label{RPderiv}
  \deriv{}{\theta}\expectw{\p{x;\theta}}{\phi(x)} &=
  \deriv{}{\theta}\expectw{\epsilon\sim\mathcal{N}(0,1)}{\phi\left(g(\epsilon;\theta)\right)}\\
  &=
  \expectw{\epsilon\sim\mathcal{N}(0,1)}{\deriv{\phi\left(g(\epsilon;\theta)\right)}{\theta}}\\
  &=
  \expectw{\epsilon\sim\mathcal{N}(0,1)}{
    \deriv{\phi\left(g(\epsilon;\theta)\right)}{g}\deriv{g(\epsilon;\theta)}{\theta}},
\end{aligned}
\end{equation}
where $\theta=[\mu,\sigma]$, $\deriv{g}{\mu} = 1$, $\deriv{g}{\sigma} = \epsilon$ and
$\deriv{\phi\left(g(\epsilon;\theta)\right)}{g} = \deriv{\phi\left(x\right)}{x}$.


What do these derivations mean, and what is the relationship between
the two methods?  We give two possible answers to this question: (i)
we give a first principles explanation that these are different methods of keeping
track of the movement of probability mass (Sec.~\ref{boxtheor}),
(ii) we show that RP and LR are \emph{duals}
under the divergence theorem when considering the integral of a probability
mass flow (Sec.~\ref{flowtheor}).
Our theory gives a physical insight by analogy to fluid dynamics, and allows for
intuitive visualizations. Our main technical result is in Thm.~\ref{flowuniquemain},
where we formalize a generalized estimator that includes all previous LR
and RP gradients as special cases, and we prove that there cannot exist an
estimator of this type outside our characterized space. Finally, we advocate
for a systematic approach in the search for novel gradient estimators (Sec.~\ref{benefit}).

\section{PRELIMINARIES}
\newtheorem*{problem}{Problem statement}

We introduce some preliminaries. In Eq.~(\ref{esticlass}) we introduce a general form of all gradient
estimators of the LR--RP type. Sec.~\ref{coordinatetrans} explains
that the introduced equation indeed includes RP as a special case.
Sec.~\ref{mcintegration} introduces the Monte Carlo (MC) integration
principle, which provides a link between integral expressions and the
corresponding gradient estimators.
Sec.~\ref{previousworks} introduces previous works on the relationship
between LR and RP, and the limitations of these works. Sec.~\ref{veccalcsec}
gives basic knowledge about fluid dynamics and the divergence theorem
necessary for understanding Sec.~\ref{flowtheor}.

\subsection{Setup}

\begin{problem}{}
  Given one sample $\bfv{x}\sim q(\bfv{x})$, and while being
  allowed to evaluate $\phi(\bfv{x})$ and
  $\nabla_{\bfv{x}}\phi(\bfv{x})$, construct an estimator, $E_{\theta_i}$, which
  may depend on
  $\bfv{x}, \phi(\bfv{x})$ and $\nabla_{\bfv{x}}\phi(\bfv{x})$, s.t.\
  \begin{equation}
    \expectw{q(\bfv{x})}{E_{\theta_i}} = \deriv{}{\theta_i}\expectw{p(\bfv{x})}{\phi(\bfv{x})}.
  \end{equation}
\end{problem}
To obtain an estimator for the full gradient w.r.t.\ $\theta$ (as
opposed to the derivative w.r.t.\ one element of the parameters
$\theta_i$), we can stack the estimators together:
$E_\theta = [E_{\theta_1}, E_{\theta_2},\ldots]$. Note that, while in the problem statement
we consider one sample, $\bfv{x}\sim q(\bfv{x})$, the estimators can also be used with multiple
samples in a batch by averaging the estimates together.  The main reason we
explicitly write out this problem statement is to emphasize the format
based on having access to $\phi(\bfv{x})$ and
$\nabla_{\bfv{x}}\phi(\bfv{x})$.  We will further restrict our
discussion to estimators having the following product form:
\begin{equation}
\label{esticlass}
E_{\theta_i} = \bfv{u}_{\theta_i}(\bfv{x})\cdot\nabla_{\bfv{x}}\phi(\bfv{x}) + \psi_{\theta_i}(\bfv{x})\phi(\bfv{x}),
\end{equation}
where $\bfv{u}_{\theta_i}(\bfv{x})$ is an arbitrary vector field, and
$\psi_{\theta_i}(\bfv{x})$ is an arbitrary scalar field (a function).
Essentially,
this equation is taking a weighted sum of the partial derivatives and value of
$\phi(\bfv{x})$, with a different weighting defined at each $\bfv{x}$.
Both LR and RP belong to this class of gradient estimators. Assuming that
the sampling distribution is $q(\bfv{x}) = p(\bfv{x};\theta)$, we obtain LR by setting
$\bfv{u}_{\theta_i}(\bfv{x}) = \bfv{0}$, and $\psi_{\theta_i}(\bfv{x}) =
\deriv{\log\p{\bfv{x};\theta}}{\theta_i}$.
Note, that in Eq.~(\ref{RPderiv}), RP also has a $\deriv{\phi(x)}{x}$ term,
so it seems that it may also be described by the class of estimators in
Eq.~(\ref{esticlass}); however, there is a coordinate transformation to the
$\epsilon$-space, which may cause some confusion. In Sec.~\ref{coordinatetrans}, we clear
this confusion and show that, indeed, RP also belongs to the given class of estimators.

\subsection{Coordinate Transformations}
\label{coordinatetrans}
In Eq.~(\ref{RPderiv}),
$\deriv{\phi\left(g(\epsilon;\theta)\right)}{g} =
\deriv{\phi\left(x\right)}{x}$ requires no reference to $\epsilon$,
and could be computed by just knowing the $x$ corresponding to the
$\epsilon$.  Moreover, each $\epsilon$ is always in a one-to-one
correspondence with a particular $x$.\footnote{Here, we
  assume that $g$ is invertible as is usually the case;
  however, this assumption can be easily lifted by integrating
  across the pre-image of $x$ (App.~\ref{rpNotUnique}).} Therefore, the whole estimator
$\deriv{\phi\left(g(\epsilon;\theta)\right)}{g}\deriv{g(\epsilon;\theta)}{\theta}$
could be computed by directly sampling $x\sim p(x;\theta)$, converting
it to the corresponding $\epsilon$, and computing the estimator. We
denote the mapping from $x$ to $\epsilon$ with
$\epsilon = S(x;\theta)$, and call $S$ the \emph{standardization
  function}. This function is the inverse of the RP transformation,
$x = g(\epsilon;\theta)$, defined in the introduction. The
standardization function was used in implicit reparameterization
gradients \citep{figurnov2018implicitRP} to create an estimator
without reference to $\epsilon$ that is applicable to a broader class
of distributions than typical reparameterizations.  The main point we
wanted to emphasize here is that there is no need to refer to
coordinate transformations at all to define RP gradients, and the
$\epsilon$-space is just a convenience that makes it easier to apply
RP using automatic differentiation. In particular, given a sample
$x\sim p(x;\theta)$, the RP estimator can be written as
$\deriv{\phi(x)}{x}\pderiv{g(\epsilon;\theta)}{\theta}{\epsilon=S(x;\theta)}$,
where $\pderiv{g(\epsilon;\theta)}{\theta_i}{\epsilon=S(x;\theta)}$ corresponds
to $\bfv{u}_{\theta_i}(\bfv{x})$.

\subsection{Importance Sampling/MC Integration}
\label{mcintegration}
\newtheorem{definition}{Definition}
\begin{definition}[Integral expression]
  An integral expression is denoted by $\int_\Omega f(x) \textup{d}x$, and
  it comprises the domain of integration $\Omega$, the function $f(x)$, and the measure of integration corresponding to
  $\textup{d}x$.
\end{definition}
The reason we make such a seemingly trivial definition is to distinguish
between the \emph{integral expression} and the \emph{value} of the integral. For
example, the integral expressions corresponding to the LR gradient (Eq.~(\ref{LRderiv}))
and the RP gradient (Eq.~(\ref{RPderiv})) are different, but they \emph{evaluate} to the
same quantity. Thus, there is a duality between the integrals; however, it is not
clear how this duality arises. In Sec.~\ref{flowtheor} we explain that the integrals are
duals under the divergence theorem. Next, we explain how these integral expressions
are related to the gradient estimators via importance sampling.

\paragraph{MC integration:} Any integral $\int f(x) ~\textup{d}x$ can
be estimated using Monte Carlo integration as follows:
\begin{equation}
  \label{mceq}
  \begin{aligned}
  \int f(x) ~\textup{d}x &= \int q(x)\frac{f(x)}{q(x)} \textup{d}x\\
&= \expectw{q(x)}{\frac{f(x)}{q(x)}},
\end{aligned}
\end{equation}
where we are importance sampling from $x\sim q(x)$.
From this method, we see that the LR gradient estimator
$E_\mathrm{LR} = \left.\deriv{p(x;\theta)}{\theta}\phi(x)\middle/q(x)\right.$ arises by
applying the MC integration principle to the integral expression
$I_\mathrm{LR} = \int \deriv{p(x;\theta)}{\theta}\phi(x)\textup{d}x$, when $q(x) = p(x;\theta)$.
Moreover, the integral expression corresponding to RP is given by
$I_\mathrm{RP} = \int p(x;\theta)\deriv{\phi(x)}{x}\pderiv{g(\epsilon;\theta)}{\theta}{\epsilon=S(x;\theta)}
\textup{d}x$, and the gradient estimator for general $q(x)$ is
$E_\mathrm{RP}~=~\frac{p(x;\theta)}{q(x)}\deriv{\phi(x)}{x}\pderiv{g(\epsilon;\theta)}{\theta}{\epsilon=S(x;\theta)}$. In general, given an integral expression for a gradient estimator,
one can always construct the estimator by directly applying MC integration.
But the reverse is also true---given an estimator, $E$, one can
always construct the integral expression as
$I = \int E \textup{d}q$, where the $\textup{d}q$ indicates that we are integrating
w.r.t.\ the measure corresponding to $q$ ($\textup{d}q$ can be considered as being
equivalent to $q(x)\textup{d}x$). \emph{Thus, there is a one-to-one correspondence
between the estimator and the integral expression, given the sampling distribution,
$q(x)$.} Previously, in machine learning (ML), importance sampling was suggested as
a principle for LR \citep{jie2010connection}, but the link to RP
has not been discussed in ML. We on the other hand, suggest importance sampling as
a key component of any gradient estimator, including RP.

\subsection{Prior Work on the Relationship between LR and RP
  Gradient Estimators}
\label{previousworks}
\paragraph{Measure theoretic view:}
LR and RP gradients are well-studied in operations research
\citep{l1991overview}, where their relationship has been described in terms
of measure theory \citep{l1990unified}. They defined the problem as
finding the gradient of an expectation of a function
$\phi(\omega;\theta)$ where the expectation is taken w.r.t.\ a
probability measure $P_\theta$. Here, $\omega$ represents a sample
in this space, and, unlike our previous definition, $\phi$ may also depend
on $\theta$.
Then, by sampling w.r.t.\ a different probability measure $G$,
independent of $\theta$, on
the same space, the expectation can be written as 
\begin{equation}
  \int \phi(\omega;\theta) \textup{d}P_\theta =
  \int \phi(\omega;\theta) \deriv{P_\theta}{G}\textup{d}G = 
  \int \phi(\omega;\theta) L_\theta\textup{d}G,
\end{equation}
where $L_\theta = \deriv{P_\theta}{G}$ is the Radon-Nikodym derivative,
a function $f$, s.t.\ $P_\theta(A) = \int_A f \textup{d}G$, where
$P_\theta(A)$ denotes the measure of set $A$. If $p(\omega;\theta)$ and
$q(\omega)$ are the pdf's of $P_\theta$ and $G$ respectively, then
we simply have $L_\theta = \frac{\p{\omega;\theta}}{q(\omega)}$ is the
likelihood ratio. Differentiating
w.r.t.\ $\theta$ gives
\begin{equation}
\label{lecuyereq}
  \deriv{}{\theta}\int \phi(\omega;\theta) L_\theta\textup{d}G =
  \int \deriv{\phi(\omega;\theta)}{\theta} L_\theta +
  \phi(\omega;\theta) \deriv{L_\theta}{\theta}\textup{d}G.
\end{equation}
Now, depending on how the probability space $P_\theta$ is defined, one
obtains either the likelihood ratio gradient or the reparameterization
gradient. If $\omega \coloneqq x$, then $\phi$ is independent of
$\theta$, so $\deriv{\phi(\omega;\theta)}{\theta} = 0$, and one
obtains the likelihood ratio gradient term
$\deriv{L_\theta}{\theta} = \deriv{\p{x;\theta}}{\theta}/q(x)$, which
is the same as in Eq.~(\ref{LRderiv}), except that importance sampling
from $q$ is used (set $q = p$ to get exactly the LR gradient). On the
other hand, if $\omega \coloneqq \epsilon$, and
$\phi(\epsilon;\theta)$ is defined as
$\phi\left(g(\epsilon;\theta)\right)$, then $L$ is independent of
$\theta$, and one is left with only the reparameterization gradient
term $\deriv{\phi(\epsilon;\theta)}{\theta}$, as in
Eq.~(\ref{RPderiv}).

A strength of this view is that it shows that LR and RP lie
at opposite ends of a spectrum of estimators using both
derivative and value information of $\phi$. However, the additional
intuition is still limited, as the theory does not explain how these
opposite ends are related, and how to convert between the two---the
theory only says that if one \emph{can} choose probability spaces with
specific properties, one obtains either RP or LR, but it does not
explain \emph{how} to achieve the desired properties.

\paragraph{Stein's identity/integration by parts:}
Another work on policy gradients \citep{liu2017action} showed a
connection between RP and LR via Stein's identity:
\begin{equation}
\label{stein}
  \int \p{x;\theta}\left(\deriv{\log\p{x;\theta}}{x}\phi(x) +
    \deriv{\phi(x)}{x}\right)\textup{d}x = 0;
\end{equation}
however, note that the derivative here is w.r.t.\ $x$, not
$\theta$. They showed algebraically that it generalizes to
derivatives w.r.t.\ $\theta$, but to do so, they put infinitesimal
Gaussian noise on $x$, and the additional intuition from their work is
still limited.

\cite{ranganath2016hierarchical} presented a derivation based on integration
by parts, which can be seen as a generalized view compared to Stein's identity.
We present this derivation and discuss it below.

Integration by parts is described by the identity:
\begin{equation}
\begin{aligned}
  &\int_a^b f(x)h(x)\textup{d}x \\ &= \left[\int_a^xf(z)\textup{d}z ~h(x)\right]_a^b
  - \int_a^b\int_a^xf(z)\textup{d}z~\deriv{h(x)}{x}\textup{d}x.
\end{aligned}
\end{equation}
We apply this identity on the integral for LR:
\begin{equation}
\label{intpartderiv}
\begin{aligned}
  &\int_{-\infty}^{+\infty}\deriv{p(x;\theta)}{\theta}\phi(x)\textup{d}x
  = \left[\int_{-\infty}^x\deriv{p(z;\theta)}{\theta}\textup{d}z
    ~\phi(x)\right]_{-\infty}^{+\infty}\\
  &- \int_{-\infty}^{+\infty}\int_{-\infty}^x\deriv{p(z;\theta)}{\theta}
  \textup{d}z~\deriv{\phi(x)}{x}\textup{d}x.
\end{aligned}
\end{equation}
We can simplify as follows:
\begin{equation}
  \int_{-\infty}^x\deriv{p(z;\theta)}{\theta}\textup{d}z =
  \deriv{}{\theta}\int_{-\infty}^xp(z;\theta)\textup{d}z = \deriv{Q(x;\theta)}{\theta},
\end{equation}
where $Q(x;\theta)$ is the cumulative density function.  The first
term in Eq.~(\ref{intpartderiv}) disappears because
$\deriv{Q(x;\theta)}{\theta} = 0$ at $x=\pm \infty$, and we end up
with
\begin{equation}
  \int p(x;\theta)\left(\frac{-1}{p(x;\theta)}\deriv{Q(x;\theta)}{\theta}\right)
  \deriv{\phi(x)}{x}\textup{d}x.
\end{equation}
We can see that $\frac{-1}{p(x;\theta)}\deriv{Q(x;\theta)}{\theta}$ is
equivalent to $\bfv{u}_{\theta_i}(\bfv{x})$ in Eq.~(\ref{esticlass}).
\footnote{Note that substituting $\theta = x$ leads to
$\deriv{Q(x;\theta)}{x}~=~p(x;\theta)$, and the equation becomes
Stein's identity in Eq.~(\ref{stein}), showing that integration by
parts generalizes it.} In the one-dimensional case, it turns out that this estimation method
is the same as RP; however, the theory is still limited in several
ways: (i) the derivation only considers one particular
$\bfv{u}_{\theta_i}(\bfv{x})$ as opposed to an arbitrary one, (ii) in
multiple dimensions, there are other RP gradients not conforming to
this equation \citep{jankowiak2018rpflow}, (iii) the additional
intuition from the derivation is limited---it appears to be just
another ``trick''. It was suggested that the derivation is insightful,
because the analytic computation of the zero term,
$\left[\int_{-\infty}^x\deriv{p(z;\theta)}{\theta}\textup{d}z
  ~\phi(x)\right]_{-\infty}^{+\infty} = 0$, is a reason for why RP has
lower variance than LR \citep{ranganath2016hierarchical,cong2019go}.
However, this argument is unsound because, on the contrary, adding a
negatively correlated 0-mean \emph{random} variable to the estimator,
known as a control variate, is a common technique to reduce the
variance \citep{greensmith2004controlvariates}.  Moreover, RP is not
guaranteed to have lower variance than LR. For example, \cite{pipps}
showed a practical situation where LR is $10^6$ more accurate than RP
due to chaotic dynamics in the system. Other works showed toy problems
where LR outperforms RP \citep{gal2016uncertainty,mcgradrev}. Finally,
it is unclear why analytically integrating a variable added to the
integral expression should be related to the variance to begin with
(as opposed to integrating a random variable in the \emph{estimator},
a technique known as conditioning/Rao-Blackwellization
\citep{mcbook}).

In conclusion, previous theories of the connection between RP and LR are still
limited.


\remove{What do these derivations mean, and what is the relationship between
the methods?  We give two possible answers to this question in
Secs.~\ref{boxtheor} and \ref{flowtheor}, then explain that the LR
gradient is the unique unbiased estimator that weights the function
values $\phi(x)$, and motivate importance sampling from a different distribution
$q(x)$ to reduce LR gradient
variance. Our optimal importance sampling scheme is reminiscent of the
optimal reward baseline for reducing LR gradient variance
\citep{weaver2001optimalbaseline} (App.~\ref{lrbasicsapp}), but our result is
orthogonal, and can be combined with such prior methods.}

\subsection{Vector Calculus and Fluid Dynamics}
\label{veccalcsec}
Our unified theory in Sec.~\ref{flowtheor} relies on considering a ``flow'' of
probability mass, so we give some background information.
We illustrate the background in the 3-dimensional case, but it generalizes
straightforwardly to higher dimensions.

\paragraph{Notation:}\mbox{}\\
$\bfv{F} = [F_x(x,y,z), F_y(x,y,z), F_z(x,y,z)]$ is a vector field. \\
$\phi(x,y,z)$ is a scalar field (a scalar function). \\
Divergence operator:
$\nabla\cdot \bfv{F} = \pderivw{F_x}{x} + \pderivw{F_y}{y} +
\pderivw{F_z}{z}$. \\Gradient operator:
$\nabla\phi = \left[\pderivw{\phi}{x}, \pderivw{\phi}{y},
\pderivw{\phi}{z}\right]$.

\begin{figure}
        \centering
\includegraphics[width=.4\textwidth]{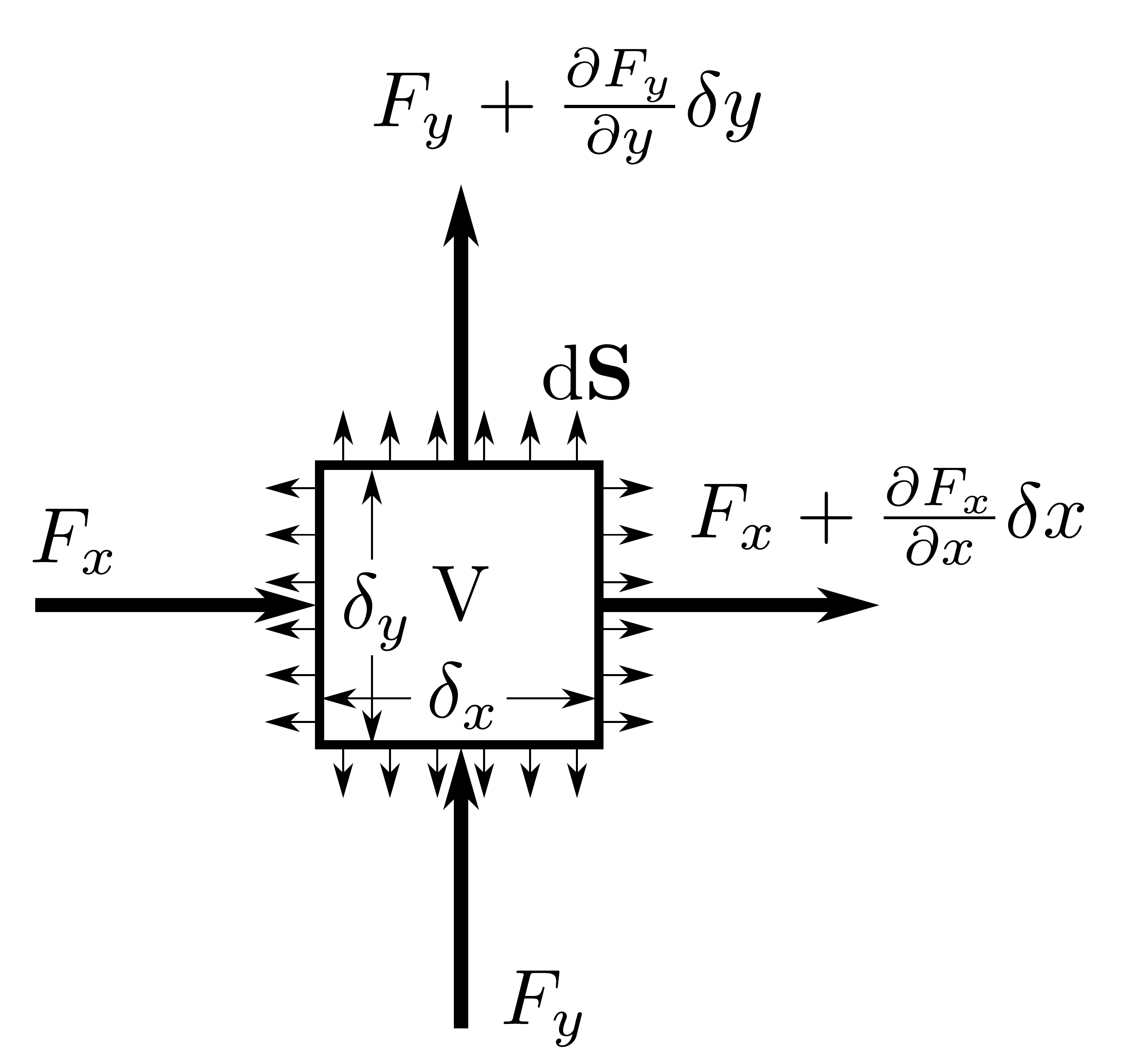}
\caption{Illustration of the divergence theorem.}
          \label{divtheoremmain}
\end{figure}

The vector field $\bfv{F}$ could be, for example, thought of as a local
flow velocity of some fluid. If $\bfv{F}$ is the density flow rate, then
the divergence operator essentially measures how much the density is decreasing at
a point. If the outflow is larger than the inflow, then the density
decreases and vice versa. The divergence theorem, illustrated in Fig.~\ref{divtheoremmain},
shows how this change in density can be measured in two equivalent ways:
one could integrate the divergence across the volume, or one could integrate the
inflow and outflow across the surface. 
\begin{figure*}[ht]
        \centering
	\begin{subfigure}{.33\textwidth}
		\includegraphics[width=\textwidth]{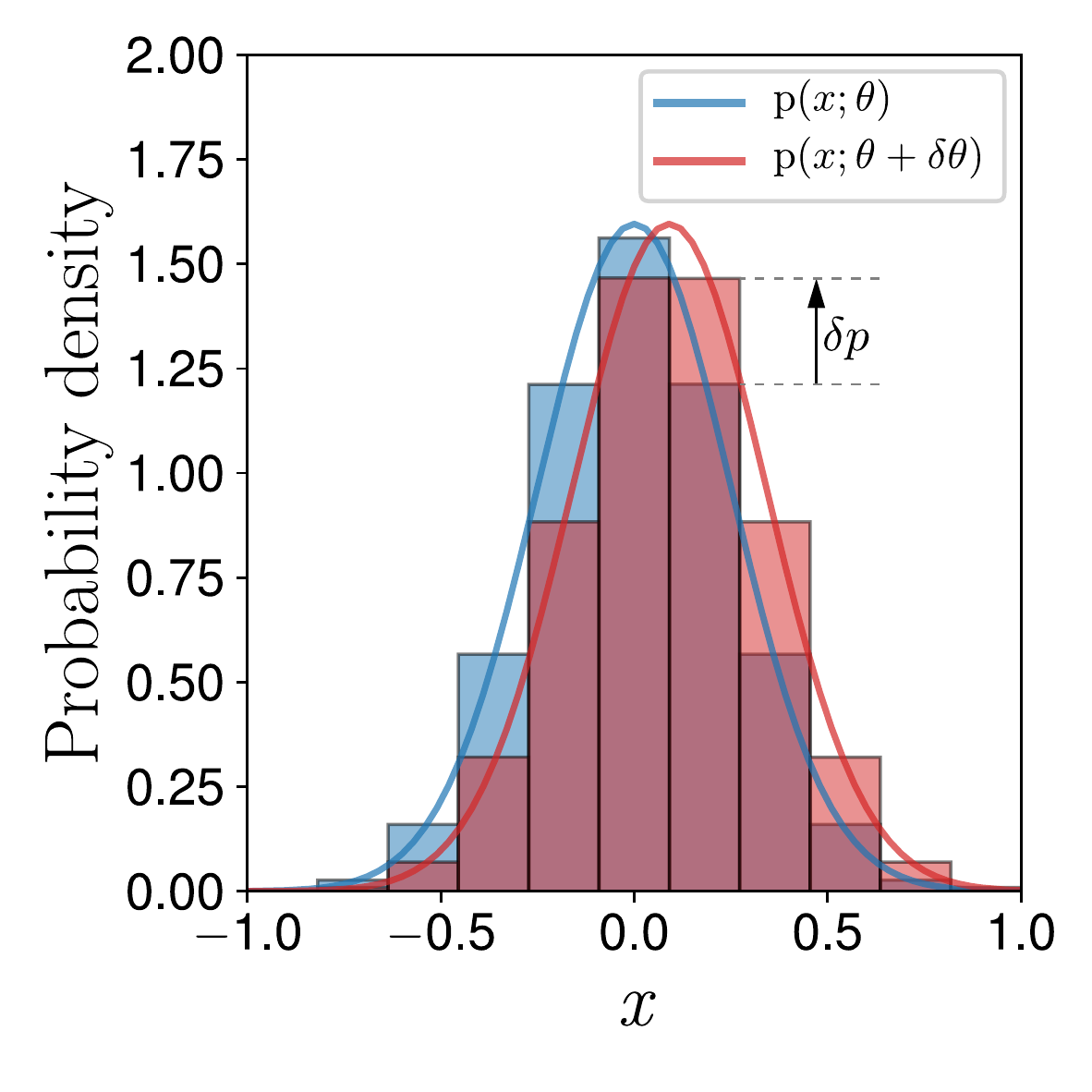}
		\caption{LR probability ``boxes''}
          \label{lrbox}
	\end{subfigure}
	\begin{subfigure}{.33\textwidth}
		\includegraphics[width=\textwidth]{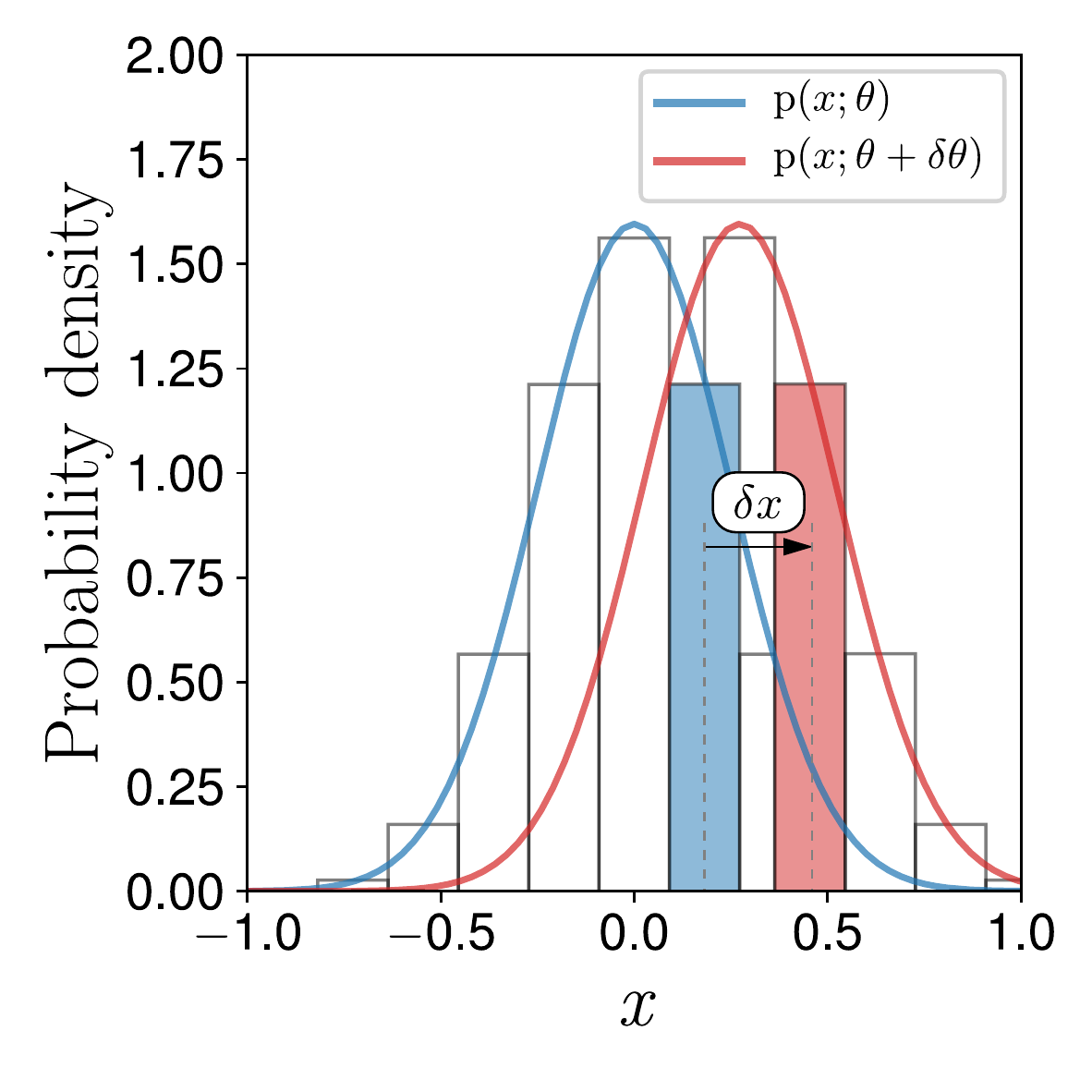}
		\caption{RP probability ``boxes''}
          \label{rpbox}
        \end{subfigure}
	\caption{LR keeps the boundaries of the ``boxes" fixed, while
          RP keeps the probability mass fixed.}
          \label{probboxes}
\end{figure*}

\begin{theorem}[Divergence theorem]
  \begin{equation}
   \label{divtheoremtop}
  \int_V \nabla\cdot\bfv{F}\textup{d}V = \int_S\bfv{F}\cdot\textup{d}\bfv{S}.
\end{equation}
\end{theorem}
\begin{proof}
To prove the claim, consider the infinitesimal box in
Fig.~\ref{divtheoremmain}.  The divergence can be calculated as
\begin{equation}
  \int_V \nabla\cdot\bfv{F}\textup{d}V =
  \delta x\delta y\left(\pderivw{F_x}{x} + \pderivw{F_y}{y}\right).
\end{equation}
On the other
hand, to take the integral across the surface, note that the surface
normals point outwards, and the integral becomes
\begin{equation}
  \begin{aligned}
    \int_S\bfv{F}\cdot\textup{d}\bfv{S} &=
    \delta_y\left(-F_x + F_x - \pderivw{F_x}{x}\delta x\right) \\
    &~~~+ \delta x\left(-F_y + F_y
    + \pderivw{F_y}{y}\delta y\right) \\
    &= \delta x\delta y\left(\pderivw{F_x}{x} +
\pderivw{F_y}{y}\right),
\end{aligned}
\end{equation}
which is the same as the divergence. To generalize this
to arbitrarily large volumes, notice that if one stacks the boxes next to each
other, then the surface integral across the area where the boxes meet cancels
out, and only the integral across the outer surface remains.
\end{proof}
For an
incompressible flow, the density at any point does not change, and the divergence
must be zero.

\section{A PROBABILITY ``BOXES" VIEW OF LR AND RP GRADIENTS}
\label{boxtheor}

Here we give our first explanation of the link between LR and RP
gradients, illustrated in Fig.~\ref{probboxes}. In short, LR gradients
estimate the change in expectation by measuring how the probability
mass assigned to each $\phi(x)$ located at a fixed $x$ changes, whereas
RP gradients define ``boxes'' of fixed probability mass, keep track of
where this ``box'' moves as the parameters $\theta$ change, and
measure how the value $\phi$ corresponding to this ``box'' changes.
For the ease of the explanation, consider a discrete space, where
$x \sim P(x;\theta)$ can take $N$ possible values. The continuous case
can be recovered by letting $N\to\infty$. Fundamentally, the
expectation, $\expectw{P(x;\theta)}{\phi(x)}$, is a weighted
average of function values $\sum_{i=1}^N P(x_i)\phi(x_i)$, where the
weights sum to one:~$\sum_{i=1}^N P(x_i) = 1$.
Therefore, to determine the expectation,
we must determine \emph{how the probability mass is allocated to the
  different $\phi(x_i)$ values}. We can envision two different
allocation procedures: (i) for each $\phi(x_i)$ at a \emph{fixed}
$x_i$, we determine how much probability mass $P_i$ we assign to it;
(ii) we predetermine the sizes of the ``boxes'' of \emph{fixed}
probability mass $P_j$, then, for each box with weight $P_j$ we assign
one of the available $\phi(x_i)$ values.  Now, to measure the gradient
of the expectation,
$\deriv{}{\theta}\expectw{P(x;\theta)}{\phi(x)}$, \emph{one must
  measure how the probability mass is reallocated} as the parameters
$\theta$ are perturbed. We will see that allocation procedure (i)
corresponds to LR gradients, whereas (ii) corresponds to RP
gradients. A full formal derivation is given in App.~\ref{probboxapp},
but reading it should not be necessary to understand the concept,
which we explain intuitively below. To perform the estimation, first
note that the gradient is given by
$\deriv{}{\theta}\sum_{i=1}^N P(x_i)\phi(x_i) = \sum_{i=1}^N
\deriv{}{\theta}\bigl(P(x_i)\phi(x_i)\bigr)$.

\paragraph{LR estimator:} In case (i): $\phi(x_i)$ is fixed, so
$\deriv{}{\theta}\bigl(P(x_i)\phi(x_i)\bigr) =
\deriv{P(x_i)}{\theta}\phi(x_i)$, and to estimate the gradient,
\emph{one must measure how the weight assigned to each particular
  $\phi(x_i)$ changes}. This corresponds precisely to what the LR
gradient estimator does. To see this, first consider that any integral
can be estimated by importance sampling from a distribution $q(x)$,
and using MC integration, as shown in Eq.~(\ref{mceq}).
Now, we set
$q(x) = P(x;\theta)$, sample $x_i \sim P(x;\theta)$, and use the
gradient estimator
$E = \frac{1}{P(x_i;\theta)}\deriv{P(x_i)}{\theta}\phi(x_i)$. Then this
will satisfy
$\expectw{x_i\sim P(x;\theta)}{E} = \sum_{i=1}^N
\deriv{}{\theta}\bigl(P(x_i)\phi(x_i)\bigr)$. Note that 
$\frac{1}{P(x_i;\theta)}\deriv{P(x_i;\theta)}{\theta} =
\deriv{}{\theta}\log P(x_i;\theta),$ and we see that it is the same
as the LR gradient in Eq.~(\ref{LRderiv}). The transformation to the
log term is known as the log-derivative trick, and
it may appear to be the essence behind the
LR gradient. However, actually the multiplication and division by
$P(x;\theta)$ is just a special case of the more general MC
integration principle. Rather than thinking of the
LR gradient in terms of the log-derivative term, it may be better
to think of it as simply estimating the integral of the probability
gradient by applying the
appropriate importance weights.
Sometimes, the LR gradient is described as being ``kind of like a
finite difference gradient"
\citep{salimans2017oaies,mania2018simplers}, but here we see that it is
a different concept that does not rely on fitting a straight line
between differences of $\phi$ (App.~\ref{lrbasicsapp}), but estimates
how probability mass is reallocated among different $\phi$ values.

\paragraph{RP estimator:} In case (ii): $P(x_i)$ is fixed, but
$\phi(x_i)$ may change---such a situation can occur when one has a fixed
amount of probability mass $P_i$ in the ``box'', but the location,
$x_i$, changes. In this case, we have
$\deriv{}{\theta}\bigl(P(x_i)\phi(x_i)\bigr) =
P(x_i)\deriv{\phi(x_i)}{\theta} =
P(x_i)\deriv{\phi(x_i)}{x_i}\deriv{x_i}{\theta}$, and to estimate the
gradient, \emph{one must measure how the function value $\phi$ in the
  ``box'' changes}. For example, consider shifting the mean location
of a Gaussian distribution by $\delta\mu$, hence, also shifting the
location of each of the ``boxes'' by the same quantity, as depicted in
Fig.~\ref{probboxes}. The probability inside the box would stay fixed,
but the function value $\phi$ would change.  This situation
corresponds to the RP gradient in Eq.~(\ref{RPderiv}).  In this case,
the position of the ``box'' is defined by
$x_i := g(\epsilon_i;\theta)$, and the probability density assigned to
$\epsilon_i$ stays fixed at $p(\epsilon_i)$. Finally, note that
we can construct an estimator
$E = \deriv{\phi(x_i)}{x_i}\deriv{x_i}{\theta}$ by sampling from
$x_i\sim P(x_i)$, and this will be unbiased:
$\expectw{x_i\sim P(x;\theta)}{E} = \sum_{i=1}^N
\deriv{}{\theta}\bigl(P(x_i)\phi(x_i)\bigr)$.

We see that LR and RP are estimating the same quantity; the difference
lies just in the way how one keeps track of the movement of the
probability mass: LR measures how the probability mass assigned to a
fixed location $x_i$ changes, whereas RP measures how the function value
$\phi$ corresponding to a moving ``box'' of probability mass changes.

\remove{
We can envision two separate ways to measure how the probability mass is
reallocated among different $\phi(x_i)$ values located at different $x_i$:
(i) one could estimate how the probability mass allocated to each $\phi(x_i)$
changes; (ii) one could index a ``box'' of probability mass $\Delta p_i$
with respect to its position on the probability distribution, measure where
this ``box'' is moving, and estimate the change in the function value
$\phi(x_i)$ corresponding to this ``box'' of probability mass. We have
illustrated these two views in Fig.~\ref{probboxes}.
}

\section{A UNIFIED PROBABILITY FLOW VIEW OF
  LR AND RP GRADIENTS}
\label{flowtheor}

\begin{figure*}[ht]
        \centering
	\begin{subfigure}{.33\textwidth}
		\includegraphics[width=\textwidth]{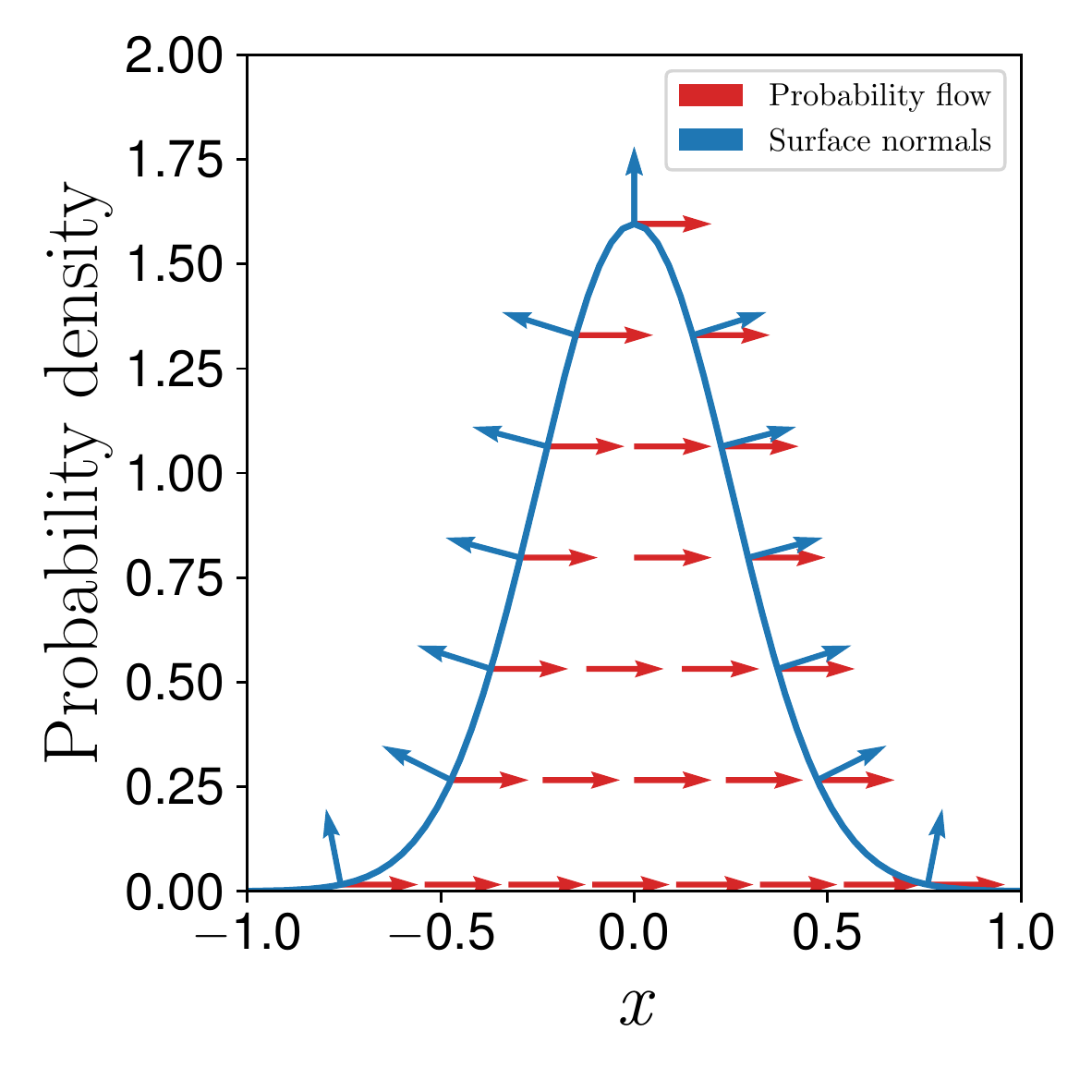}
		\caption{$\mu$ flow lines}
          \label{mulines}
        \end{subfigure}
	\begin{subfigure}{.33\textwidth}
		\includegraphics[width=\textwidth]{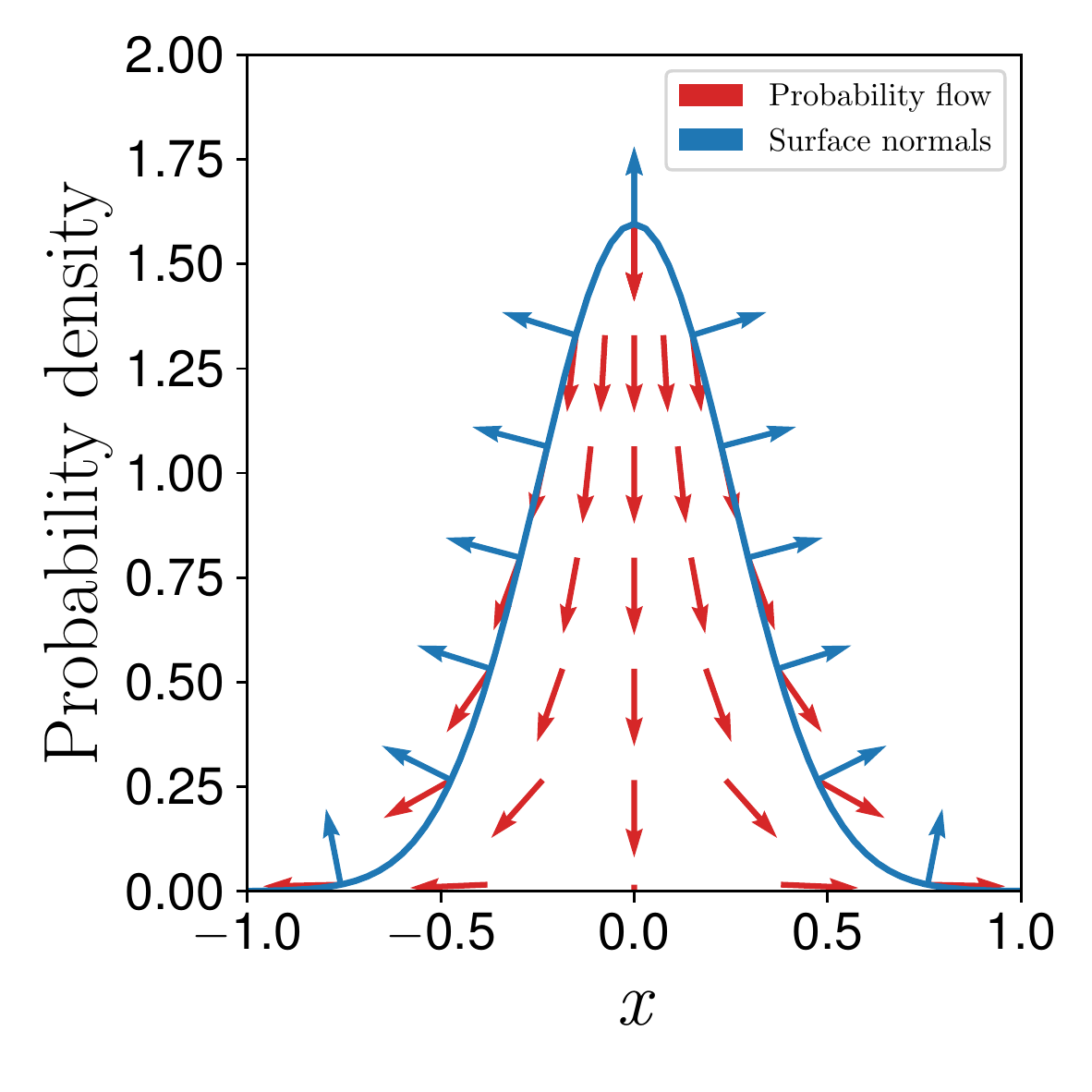}
		\caption{$\sigma$ flow lines}
          \label{sigmalines}
	\end{subfigure}
	\caption{Probability flow lines when $\mu$ and $\sigma$ are perturbed.}
          \label{problines}
\end{figure*}

Here we give another explanation of LR and RP.
In this theory, both LR and RP come out of the
same derivation, thus showing a link between the two. In particular,
we define an incompressible flow of probability mass imposed by
perturbing the parameters $\theta$ of $\p{x;\theta}$, which can
be used to express the derivative of the expectation as an integral
over this flow.
LR and RP estimators correspond to duals of this
integral under the well-known divergence theorem (Thm.~\ref{divtheoremmain}).

The main idea resembles RP, but in addition to sampling
$\bfv{x}$, we sample a height $h$ for each point:
$h~=~\epsilon_h\p{\bfv{x};\theta}$, where
$\epsilon_h \sim \textup{unif}(0,1)$, i.e., the sampling space is
extended with an additional dimension for the height
$\tilde{\bfv{x}} := [\bfv{x}^T, h]^T$, and we are uniformly sampling in
the volume under $\p{\bfv{x};\theta}$. The
definition of $g$ in the introduction is extended, s.t.\
$\tilde{g}(\epsilon_x,\epsilon_h) := \tilde{\bfv{x}} = [g(\epsilon_x)^T, \epsilon_h\p{\bfv{x};\theta}]^T$.
The expectation
turns into
\begin{equation}
\begin{aligned}
  &\deriv{}{\theta}\int\p{\bfv{x};\theta}\phi(\bfv{x})\textup{d}\bfv{x}\\
   &= \deriv{}{\theta}\int_{\epsilon_x}\int_{\epsilon_h}
  \p{\epsilon_x}\p{\epsilon_h}\phi\left(\tilde{g}(\epsilon_x,\epsilon_h)\right)
    \textup{d}\epsilon_x\textup{d}\epsilon_h\\
    &=   \int_{V}\nabla_{\theta}\phi(\tilde{g}(\epsilon_x,\epsilon_h))
    \textup{d}V
    =
  \int_{V}\nabla_{\tilde{\bfv{x}}}\phi(
  \tilde{\bfv{x}})\nabla_\theta \tilde{g}(\epsilon_x,\epsilon_h)
    \textup{d}V.
  \end{aligned}
\label{volint}
\end{equation}
In Eq.~(\ref{volint}), $V$ is the volume under the curve of $p(\bfv{x};\theta)$, and
$\phi([\bfv{x}^T, h]^T) \coloneqq \phi(\bfv{x})$ ignores the
$h$-component. Each column $i$ of
$\nabla_\theta \tilde{g}(\epsilon_x,\epsilon_h)$ corresponds to a vector field
induced by perturbing the $i^{\textup{th}}$ component of $\theta$. The
red lines in Fig.~\ref{problines} show the flow fields for a
Gaussian distribution as the mean and variance are perturbed.  The
other term,
$\nabla_{\tilde{\bfv{x}}}\phi(\tilde{\bfv{x}})$, is the gradient of the
scalar field $\phi(\tilde{\bfv{x}})$. As $\phi$ is independent of
$h$, the gradient is parallel to the $\bfv{x}$ axes with
magnitude $\deriv{\phi}{\bfv{x}}$.

According to the divergence theorem in Eq.~(\ref{divtheoremtop}), the
\emph{volume integral} in
Eq.~(\ref{volint}) can be turned into a \emph{surface integral} over the
boundary $\bfv{S}$ ($\textup{d}\bfv{S}$ is a
shorthand for $\hat{\bfv{n}}\textup{d}S$, where $\hat{\bfv{n}}$ is the surface normal
vector), depicted by the blue lines in Fig.~\ref{problines}.
%
%

In Eq.~(\ref{divtheoremtop}), $\bfv{F}$ is any vector field. A common corollary
arises by picking $\bfv{F} = \phi\bfv{v}$, where $\phi$ is a scalar
field, and $\bfv{v}$ is a vector field. We choose
$\bfv{v} = \nabla_\theta \tilde{g}(\epsilon_x,\epsilon_h)\delta\theta$, where
$\delta\theta$ is an arbitrary perturbation in $\theta$, so that
$\bfv{F} = \phi(\tilde{\bfv{x}})\nabla_\theta
\tilde{g}(\epsilon_x,\epsilon_h)\delta\theta$, in which case
$\nabla_{\tilde{\bfv{x}}}\cdot\bfv{F} =
\nabla_{\tilde{\bfv{x}}}\cdot\left(\phi(\tilde{\bfv{x}})\nabla_\theta
  \tilde{g}(\epsilon_x,\epsilon_h)\delta\theta\right) =
\nabla_{\tilde{\bfv{x}}}\phi(\tilde{\bfv{x}})\nabla_\theta
\tilde{g}(\epsilon_x,\epsilon_h)\delta\theta +
\phi(\tilde{\bfv{x}})\nabla_{\tilde{\bfv{x}}}\cdot\nabla_\theta
\tilde{g}(\epsilon_x,\epsilon_h)\delta\theta$. Note that the term
$\nabla_\theta \tilde{g}(\epsilon_x,\epsilon_h)\delta\theta$ corresponds to an
incompressible flow (because the probability density does not change
at any point in the augmented space). As the divergence of an incompressible
flow is 0,
$\nabla_{\tilde{\bfv{x}}}\cdot\nabla_\theta
\tilde{g}(\epsilon_x,\epsilon_h)\delta\theta = 0$, and the second term
disappears. Noting that $\delta\theta$ can be canceled, because it is
arbitrary, we are left with the equation:
\begin{equation}
\begin{aligned}
  \int_{V}\nabla_{\tilde{\bfv{x}}}\phi(
  \tilde{\bfv{x}})\nabla_\theta \tilde{g}(\epsilon_x,\epsilon_h)
    \textup{d}V &= \int_{S} \phi(
  \tilde{\bfv{x}})\nabla_\theta \tilde{g}(\epsilon_x,\epsilon_h)\cdot\textup{d}\bfv{S}.
  \end{aligned}
\label{surfint}
\end{equation}
Now we explain how the left-hand side of Eq.~(\ref{surfint}) gives rise
to the RP gradient estimator, while the right-hand side corresponds to
the LR gradient estimator. \remove{The RP estimator follows quite easily, while the
LR estimator is slightly more involved.}

\paragraph{RP estimator:}
Consider the
$\nabla_{\tilde{\bfv{x}}}\phi( \tilde{\bfv{x}})\nabla_\theta
\tilde{g}(\epsilon_x,\epsilon_h)$ term. As the scalar field
$\phi( \tilde{\bfv{x}})$ is independent of the height location $h$,
the component of the gradient in that direction is 0, and
$\nabla_{\tilde{\bfv{x}}}\phi(\tilde{\bfv{x}}) =
[\nabla_{\bfv{x}}\phi(\bfv{x}), 0]$.
As the $h$-component is 0,
the value of $\tilde{g}$ in the $h$-direction is multiplied by 0, and is
irrelevant for the product, so
$\nabla_{\tilde{\bfv{x}}}\phi( \tilde{\bfv{x}})\nabla_\theta
\tilde{g}(\epsilon_x,\epsilon_h) = \nabla_{\bfv{x}}\phi(\bfv{x}) \nabla_\theta
g(\epsilon_x)$, which is just the term used in the RP
estimator. Hence, the left-hand side of Eq.~(\ref{surfint}) corresponds
to the RP gradient.

\paragraph{LR estimator:} We will show that the LR estimator tries to
integrate
$\int_S\phi( \tilde{\bfv{x}})\nabla_\theta
\tilde{g}(\epsilon_x,\epsilon_h)\cdot\textup{d}\bfv{S}$. To do so, note that
$\textup{d}\bfv{S} = \hat{\bfv{n}}\textup{d}S$. It is necessary to
express the normalized surface vector $\hat{\bfv{n}}$, and then perform
the integral over the surface. The derivation is in App.~\ref{surfDeriv},
and the final result is
\begin{equation}
\begin{aligned}
  \int_{S} \phi(\tilde{\bfv{x}})\nabla_\theta
  \tilde{g}(\epsilon_x,\epsilon_h)\cdot\textup{d}\bfv{S} =
  \int_{X}
  \phi(\bfv{x})\deriv{\p{\bfv{x};\theta}}{\theta}
  ~\textup{d}\bfv{x}.
\label{eq:subs}
\end{aligned}
\end{equation}
We have already seen that MC integration of
the right-hand side of Eq.~(\ref{eq:subs}) using samples from
$\p{\bfv{x};\theta}$ yields the LR
estimator. Thus, RP and LR
are duals under the divergence theorem. To further strengthen this
claim we prove that the LR gradient estimator is the unique estimator
that takes weighted averages of the function values $\phi(\bfv{x})$.
\begin{theorem}[Uniqueness of LR estimator]
\label{lrunique}
\mbox{}\\$\psi(\bfv{x}) = \p{\bfv{x};\theta}\deriv{\log \p{\bfv{x};\theta}}{\theta}$ is
  the unique function $\psi$, s.t.\
  $\int \psi(\bfv{x})\phi(\bfv{x})\textup{d}\bfv{x} =
  \deriv{}{\theta}\int\p{\bfv{x};\theta}\phi(\bfv{x})\textup{d}\bfv{x}$
  for all $\phi$.
\begin{proof}
  Suppose that there exist $\psi \neq f$,
    s.t.\
  $\int \phi(\bfv{x})\psi(\bfv{x})~\textup{d}\bfv{x} = \int
  \phi(\bfv{x})f(\bfv{x})~\textup{d}\bfv{x}$ for all $\phi$.
  Rearrange the equation into
  $\int
  \phi(\bfv{x})\left(\psi(\bfv{x})-f(\bfv{x})\right)~\textup{d}\bfv{x} =
  0$, then pick $\phi(\bfv{x}) = \psi(\bfv{x})-f(\bfv{x})$ from which we
  get
  $\int \left(\psi(\bfv{x})-f(\bfv{x})\right)^2~\textup{d}\bfv{x} = 0$.
  This leads to $\psi = f$, which is a contradiction. Therefore, there
  cannot exist such $\psi \neq f$ that satisfy the condition
  for all $\phi$.
\end{proof}
\end{theorem}
The result also follows from the Riesz representation theorem
\citep{riesz1907espece}.  From the theorem, we see that Eq.~(\ref{eq:subs}) was
immediately clear without having to go through the derivation in
App.~\ref{surfDeriv}.
\remove{(given that the RP gradient estimator is
  unbiased).} The same analysis does not work for RP
(App.~\ref{rpNotUnique}). Indeed, there are infinitely many RP
gradients \citep{jankowiak2018rpflow}. \remove{Moreover, the analysis does not
consider gradient estimators with coupled samples
\citep{walder2019newtricks,mcgradrev}, such as measure valued derivatives.}
\paragraph{Characterizing the space of all LR and RP estimators:}
Now we can derive a \emph{concrete} form of all estimators in the
\emph{abstract} class in Eqs.~(\ref{esticlass})~and~(\ref{lecuyereq}). The flow
theory assumed that the flow is aligned with the change of the
probability density. We can lift this restriction by subtracting the
excess probability mass (App.~\ref{flowgradproofs}), giving a
general gradient estimator combining both $\phi(\bfv{x})$ and
$\nabla_{\bfv{x}}\phi(\bfv{x})$. This characterization is formalized
in the theorem below. 

  \begin{theorem}[The probability flow gradient estimator characterizes the space
  of all LR--RP gradient estimators]
  Given a sample, $\bfv{x}\sim q(\bfv{x})$, every unbiased gradient estimator, $E_{\theta_i}$,
  s.t.\
  $\expectw{q(\bfv{x})}{E_{\theta_i}} = \deriv{}{\theta_i}\expectw{p(\bfv{x};\theta)}{\phi(\bfv{x})}$, of the product
  form
  in Eq.~(\ref{esticlass}),
  \[E_{\theta_i}
    = \bfv{v}(\bfv{x})\cdot\nabla_{\bfv{x}}\phi(\bfv{x}) +
      \psi(\bfv{x})\phi(\bfv{x}),\] where $\phi$ is an
      arbitrary function, and assuming\footnote{ Note that the
  case where
  $\p{\bfv{x};\theta}\phi(\bfv{x})\bfv{v}(\bfv{x})\not\to                                                                        
  0$ does not correspond to any sensible estimator, because the value
  of $\phi(\bfv{x})$ at $\lVert\bfv{x}\rVert\to\infty$ will
  influence the gradient estimation. In that case,
  because $\p{\bfv{x};\theta}\to 0$ the probability of sampling at
  infinity will tend to $0$, and the gradient variance will explode. This condition does however mean that if one wants to construct
  a sensible estimator, care must be taken to ensure that
  $\bfv{u}_{\theta_i}(\bfv{x})$ does not go to infinity too fast, e.g., as
explained by \citet{jankowiak2019pathmulti}.} $p(\bfv{x};\theta)\bfv{v}(\bfv{x})\phi(\bfv{x}) \to 0$ as $\lVert x\rVert\to\infty$,
      is a special case of the estimator characterized
  by
  \begin{equation}\begin{aligned}
      &E_{\theta_i} = \frac{\p{\bfv{x};\theta}}{q\left(\bfv{x}\right)}
    \bfv{u}_{\theta_i}(\bfv{x})\cdot\nabla_{\bfv{x}}\phi(\bfv{x})
    \\
    &+ \frac{1}{q\left(\bfv{x}\right)}\biggl(\nabla_{\bfv{x}}\cdot
  \bigl(\p{\bfv{x};\theta}\bfv{u}_{\theta_i}(\bfv{x})\bigr)
  + \deriv{\p{\bfv{x};\theta}}{\theta_i}\biggr)\phi(\bfv{x}),
    \end{aligned}
    \label{probflowgrad}
\end{equation}
  where $\bfv{u}_{\theta_i}$ is an arbitrary vector field. Note that, for simplicity, we also assumed continuity of
  $p$, $\phi$ and $\bfv{u}_{\theta_i}$; however, this is unnecessary, and discontinuities are
  handled in App.~\ref{flowdiscontapp}.
  \label{flowuniquemain}
\end{theorem}
\begin{proof}It is analogous to Thm~\ref{lrunique}. See App.~\ref{flowgradproofs}.\end{proof}

The theorem says that given $\bfv{v}$, one can derive the unique $\psi$
  necessary for unbiasedness. 
Thus, the probability flow gradient in Eq.~(\ref{probflowgrad}) generalizes
all previous LR--RP gradients in the literature, as well as all
possible gradient estimators having the product form in
Eq.~(\ref{esticlass}).\footnote{Note, there still exist other gradient
  estimators that do not have the form
  $\bfv{u}(\bfv{x})\cdot\nabla_{\bfv{x}}\phi(\bfv{x})+\psi(\bfv{x})
  \phi(\bfv{x})$, e.g., gradient estimators with coupled samples
  \citep{walder2019newtricks,mcgradrev}, or gradient estimators using
  $\frac{\textup{d}^2\phi(\bfv{x})\hfill}{\textup{d}\bfv{x}^2\hfill}$.
} By setting $\bfv{u}_{\theta_i}(\bfv{x}) = \bfv{0}$ one recovers LR;
by setting $\nabla_{\bfv{x}}\cdot
\bigl(\p{\bfv{x};\theta}\bfv{u}_{\theta_i}(\bfv{x})\bigr) +
\deriv{\p{\bfv{x};\theta}}{\theta_i} = 0$, one recovers the pathwise
estimators described by \citet{jankowiak2018rpflow}.\footnote{ The
  work of \citet{jankowiak2018rpflow} was concurrent to our initial
  derivations, and is the most similar publication to ours. They also
  used the divergence theorem, but they focused on deriving new
  pathwise estimators, and did not discuss the duality between LR and
  RP.  } Estimators combining both $\phi(\bfv{x})$ and
$\nabla_{\bfv{x}}\phi(\bfv{x})$, such as the generalized RP gradient
\citep{ruiz2016generalizedRP}, also conform to
Eq.~(\ref{probflowgrad}) (App.~\ref{flowgradexamples}).\footnote{ Note
  that it is an open question whether the generalized RP and
  probability flow gradient estimator spaces are equal. To show that
  they are equal, one would have to find a generalized
  reparameterization corresponding to each arbitrary
  $\bfv{u}_{\theta_i}(\bfv{x})$. One main difficulty would arise if one
  requires a single reparameterization to simultaneously correspond to
  multiple different $\bfv{u}_{\theta_i}(\bfv{x})$ and
  $\bfv{u}_{\theta_j}(\bfv{x})$ for different dimensions $i$ and $j$
  of the parameter vector $\theta$. However, we believe that if finding such
  reparameterization is possible at all, the reparameterization corresponding to some complicated
  flow field may be quite bizarre, while in the flow framework, one
  just has to do a dot product between the flow and the gradient to
  compute the estimator.} Moreover, estimators in discontinuous
situations, such as the GO gradient \citep{cong2019go} or RP gradients
for discontinuous models \citep{lee2018rpnondiff} also conform to this
equation when taking into account for the discontinuities
(App.~\ref{flowdiscontapp}).

The terms in Eq.~(\ref{probflowgrad}) can be readily interpreted.
First note that $q(\bfv{x})$ is just a factor due to
MC integration by sampling from $q$ (Sec.~\ref{mcintegration}).
The remaining terms can be made analogous to fluid motion. In the analogy,
perturbing $\theta_i$ is equivalent to perturbing time measured in seconds (s),
$p(\bfv{x};\theta)$ is equivalent to the density measured in $\frac{\textup{kg}\hfill}{\textup{m}^3}$,
and $\bfv{u}_{\theta_i}$ is equivalent to the flow velocity measured in
$\frac{\textup{m}}{\textup{s}}$. The term, $p(\bfv{x};\theta)\bfv{u}_{\theta_i}(\bfv{x})$, is the probability
mass flow rate per unit area, as is clear from multiplying the units:
$\frac{\textup{kg}\hfill}{\textup{m}^3}\times \frac{\textup{m}}{\textup{s}} =
\frac{\nicefrac{\textup{kg}}{\textup{s}}\hfill}{\textup{m}^2}$. The
term $\nabla_{\bfv{x}}\cdot\bigl(\p{\bfv{x};\theta}\bfv{u}_{\theta_i}(\bfv{x})\bigr)$
is the divergence of the probability mass flow rate, and it tells us the rate of
change of density at $\bfv{x}$ with a corresponding $\phi(\bfv{x})$ caused by the probability flow,
$p(\bfv{x};\theta)\bfv{u}_{\theta_i}(\bfv{x})$
(see Sec.~\ref{veccalcsec}). The
$\p{\bfv{x};\theta}\bfv{u}_{\theta_i}(\bfv{x})\cdot\nabla_{\bfv{x}}\phi(\bfv{x})$ term,
on the other hand, gives the rate of change of the $\phi(\bfv{x})$ corresponding
to a point moving on the probability flow. Integrated across the whole volume,
the two terms involving $\bfv{u}_{\theta_i}$ cancel out, leaving only the $\deriv{p(\bfv{x};\theta)}{\theta_i}$ term.
The probability flow estimator could thus also be interpreted as adding a control variate to
the standard LR gradient estimator.

Our explanation of the principle behind LR and RP gradients improves
over previous explanations based on two main criteria:
(i) greater generality, (ii) requiring fewer assumptions. In particular,
Occam's razor states that given competing explanations, one should
prefer the one with fewer assumptions. Our explanation does not require
the reparameterization assumption used in many previous explanations of
RP gradients. Instead, we argue that reparameterization is just
a trick that allows implementing the estimator easily using
automatic differentiation, but has little to do with the
principle behind its operation. Moreover, the sufficient conditions for the probability flow
gradient estimator are also necessary, so the explanation of the
principle cannot be improved without expanding the class of estimators
to go beyond Eq.~(\ref{esticlass}).

  Our characterization showed that the flow field
  $\bfv{u}_{\theta_i}(\bfv{x})$ and importance sampling distribution
  $q(\bfv{x})$, together, fully describe the space of estimators; one
  remaining question is how to pick $\bfv{u}_{\theta_i}(\bfv{x})$ and
  $q(\bfv{x})$, s.t.\ the variance of the estimator is low.
  \citet{jankowiak2018rpflow} discussed how to pick
  $\bfv{u}_{\theta_i}(\bfv{x})$ when the $\phi(\bfv{x})$ term
  disappears (i.e., for RP). In our concurrent work \citep{parmas2019unified},
  we are discussing how
  to pick $q(\bfv{x})$ when the $\nabla_{\bfv{x}}\phi(\bfv{x})$ term
  disappears (i.e., for LR). The general question of the best
  combination of $\bfv{u}_{\theta_i}(\bfv{x})$ and $q(\bfv{x})$
  remains an open problem.
\remove{
\begin{figure*}
        \centering
	\begin{subfigure}{.244\textwidth}
          \mbox{}\\[0.8em] 
		\includegraphics[width=\textwidth]{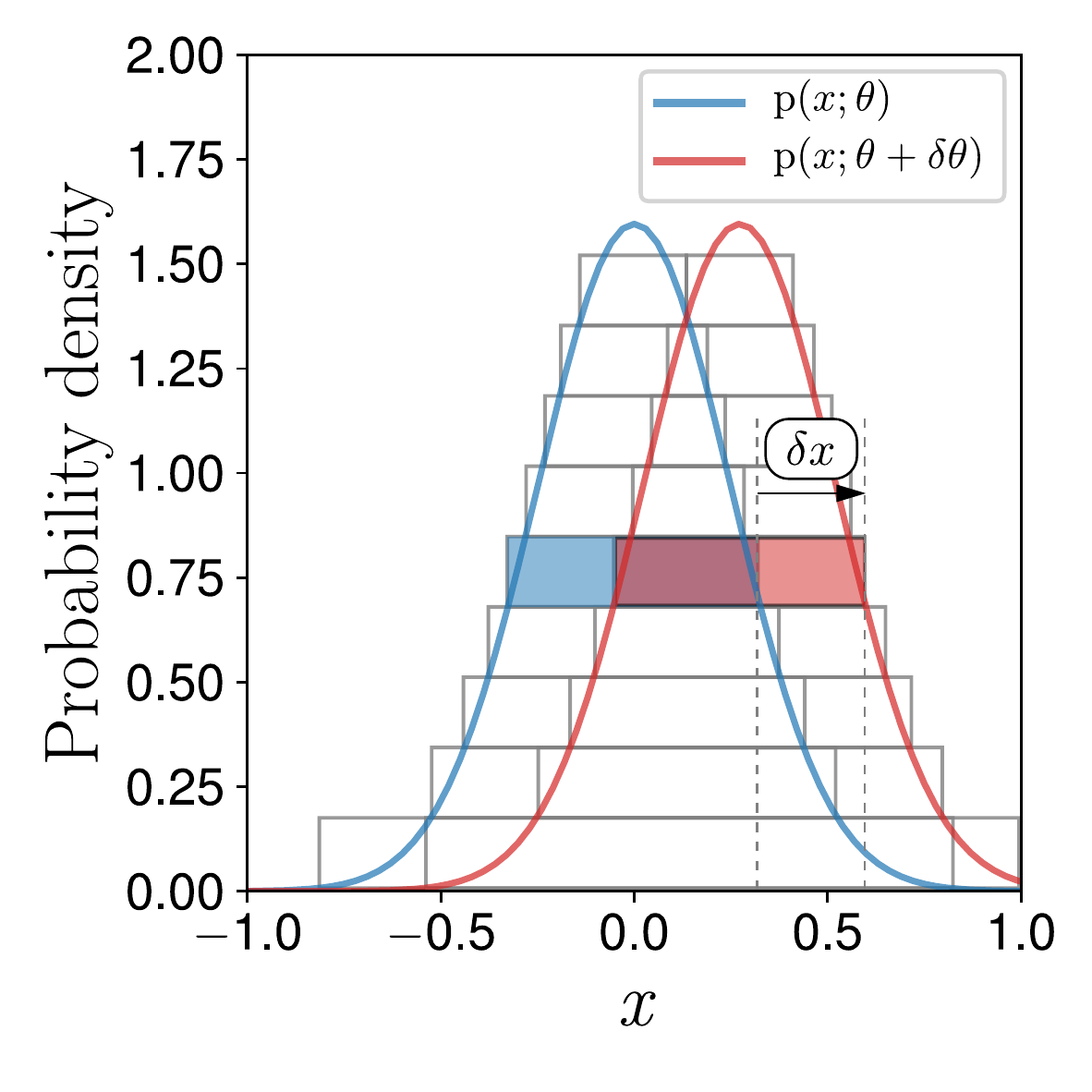}
		\caption{Horizontal probability slices}
          \label{lebbars}
        \end{subfigure}
	\begin{subfigure}{.33\textwidth}
          \includegraphics[width=\textwidth]{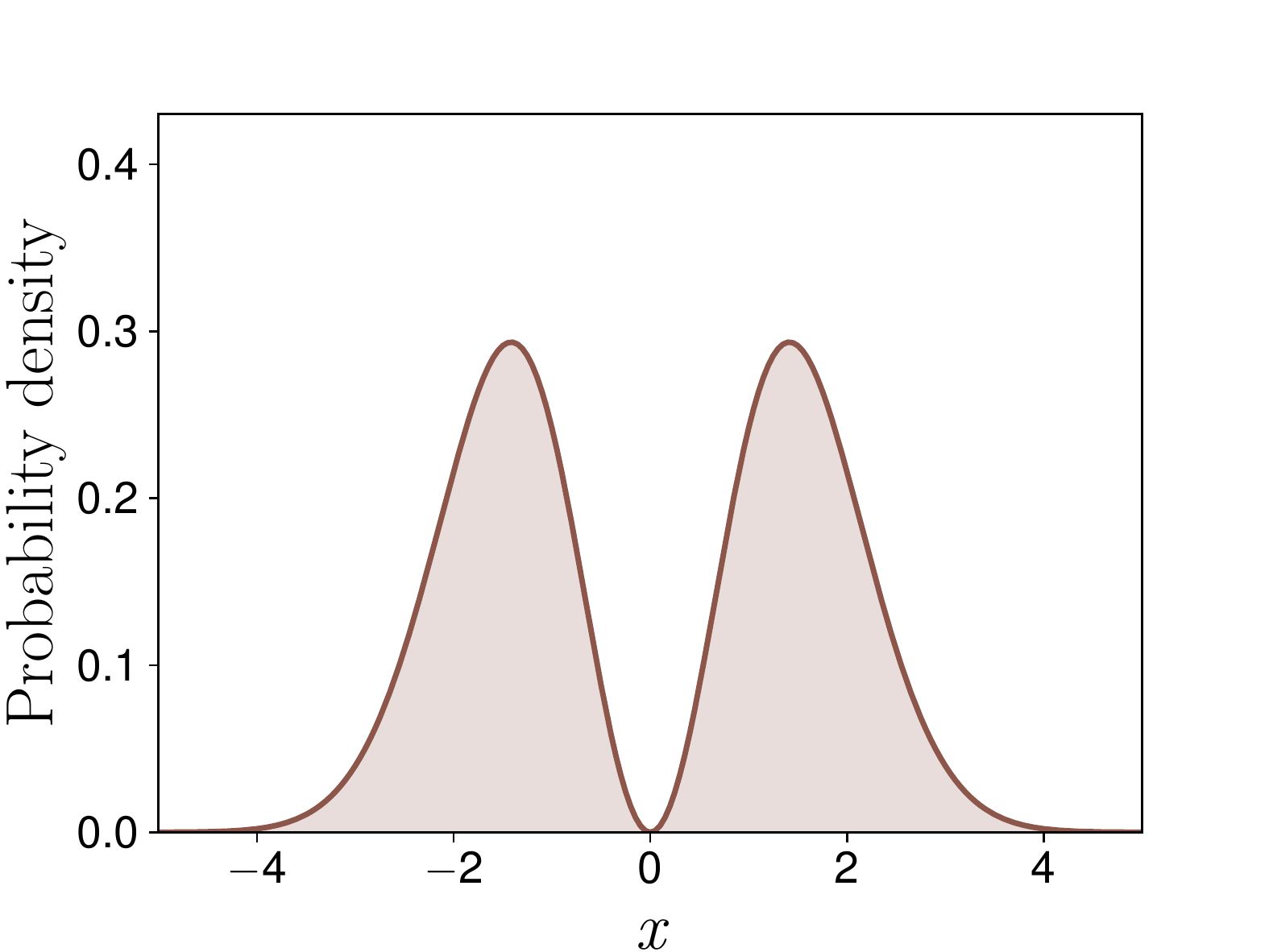}\\[-0.6em]
          \caption{Probability density}
          \label{ldist}
	\end{subfigure}
	\caption{Motivation for slice integral sampling and the distribution when the
          nominal distribution is Gaussian.}
          \label{lebbarplot}
\end{figure*}
}
\section{BENEFIT OF CHARACTERIZING THE
  SPACE OF ESTIMATORS}
\label{benefit}
\newtheorem{casestudy}{Case study}
\begin{figure*}
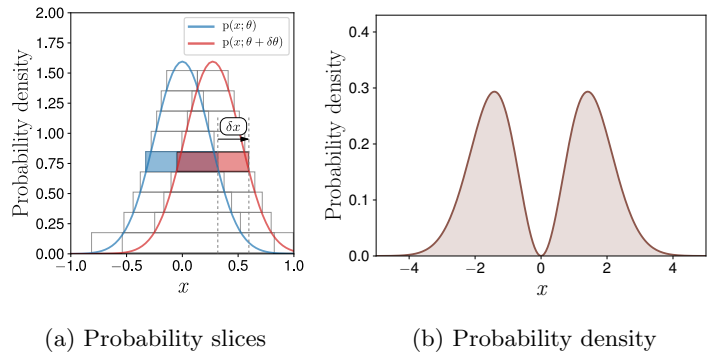

        \centering
	\begin{subfigure}{.244\textwidth}
          \mbox{}\\[0.65em] 
		\includegraphics[width=\textwidth]{LebBars3.pdf}
		\caption{Probability slices}
          \label{lebbarsmain}
        \end{subfigure}
	\begin{subfigure}{.33\textwidth}
          \includegraphics[width=\textwidth]{lebplot.pdf}\\[-0.6em]
          \caption{Probability density}
          \label{ldistmain}
	\end{subfigure}
	\caption{Motivation for slice integral sampling and the importance distribution for a
           Gaussian base distribution.}
          \label{lebbarplotmain}
\end{figure*}

Characterizing the space of estimators via uniqueness claims is
highly useful because it clarifies where we should be searching for
new gradient estimators. In particular, often a new idea might appear
promising at first sight, but uniqueness claims could immediately say that
the idea will not lead to something novel, and will instead be a special
case of the characterized space.
We illustrate this concept with two case studies below.

\begin{casestudy}[Height reparameterization]
  Consider the example with a height reparameterization explained in
  Sec.~\ref{flowtheor}. After some derivations, we reached
  Eq.~(\ref{surfint}), which we repeat below:
\begin{equation*}
\begin{aligned}
  \int_{V}\nabla_{\tilde{\bfv{x}}}\phi(
  \tilde{\bfv{x}})\nabla_\theta \tilde{g}(\epsilon_x,\epsilon_h)
    \textup{d}V &= \int_{S} \phi(
  \tilde{\bfv{x}})\nabla_\theta \tilde{g}(\epsilon_x,\epsilon_h)\cdot\textup{d}\bfv{S}.
  \end{aligned}
\end{equation*}
\end{casestudy}
This equation looks promising because the left-hand side is known to
correspond to RP, which
is an unbiased gradient estimator. Therefore, we know that the right-hand side
must also be unbiased, and it could potentially lead to some new interesting estimator
using the $\phi(\bfv{x})$ information. However, to compute the estimator, we have to
perform the tedious error-prone derivations in App.~\ref{surfDeriv}. Instead,
based on the uniqueness claim in Thm.~\ref{lrunique} we can immediately say
that this approach cannot possibly lead to a new estimator, because it
has the same form as LR (product with $\phi(\bfv{x})$), saving us the trouble of going through the derivation.

\begin{casestudy}[Horizontal slice integral sampling]
  Consider an algorithm that would sample horizontal slices of probability mass,
  illustrated in Fig.~\ref{lebbarsmain}, and motivated by flipping the vertical
  slices in Sec.~\ref{boxtheor}. One could integrate the expectation
  over the slice, and try to estimate the gradient w.r.t.\ a parameter $\theta$.
\end{casestudy}

\remove{
From Theorem~\ref{lrunique} we saw that unlike the RP gradient case,
the weighting $\psi$ for function values $\phi(\bfv{x})$ with
$\bfv{x} \sim \p{\bfv{x};\theta}$ to obtain an unbiased estimator for
the gradient $\deriv{}{\theta}\expect{\phi(\bfv{x})}$ is unique. The
only option to reduce the variance by changing the weighting would
then be to sample from a different distribution
$q(\bfv{x})$ via importance sampling. Motivated by the
resemblance of the ``boxes" theory in Sec.~\ref{boxtheor} to the
Riemann integral, we propose to sample horizontal slices of
probability mass resembling the Lebesgue integral.}

Such an approach
appears attractive, because if the location of the slice is moved by
modifying the parameters of the distribution (e.g., by changing the
mean, $\mu$), then the derivative of the expected value of the integral over
the slice will depend only on the value at the edges of the slice
(because the probability density in the middle would not change). To clarify,
consider a uniform distribution $p(x;\mu)$ between $\mu \pm \Delta$. The derivative
is $\deriv{}{\mu}\int_{\mu - \Delta}^{\mu + \Delta}p(x;\mu)\phi(x)\textup{d}x =
C\left(\phi(\mu+\Delta) - \phi(\mu-\Delta)\right)$, where $C$ is a constant. We
could use importance sampling to sample on one of the two edges of the slice, to obtain
an unbiased gradient estimator. However, at this point, it is clear that the
estimator will belong to the same class as LR, as it will be averaging some
value multiplied with $\phi(x)$, so we can already say that this idea is not
promising. At best, it would lead to an LR gradient with a different
sampling distribution $q(x)$---this is indeed the case, and the distribution
is plotted in Fig.~\ref{ldistmain}. The full derivation is in App.~\ref{sliceintegral}.

\paragraph{Systematic approach to deriving estimators:}
Rather than pursuing an ad hoc approach as in the two case studies, we propose
that we should be using a more systematic approach in the search for new
gradient estimators. It is not enough to find \emph{one} novel estimator; one should
find \emph{all} estimators of a given novel class. Our proposed 3-step approach is below.

\begin{enumerate}
\item Find a new principle for a novel gradient estimator.\mbox{}\\
\emph{In the case of LR and RP, the principle is to measure the movement
of probability mass, as described in Sec.~\ref{boxtheor}.}
\item Parameterize the class of estimators that encloses all estimators
  embodying the said principle.\\
  \emph{In our case, this parameterization is in Eq.~(\ref{esticlass}).}
\item Find necessary and sufficient conditions for the estimator to be unbiased.\\
\emph{In our case, these conditions were given in Sec.~\ref{flowtheor}.}
\end{enumerate}

\section{CONCLUSIONS}

We introduced a complete unified theory of LR and RP gradients, and
characterized the space of all unbiased single sample gradient estimators taking a weighted
sum of $\phi(\bfv{x})$ and $\nabla_{\bfv{x}}\phi(\bfv{x})$. Each
estimator is defined by a vector field
$\bfv{u}_{\theta_i}(\bfv{x})$ and an importance sampling distribution
$q(\bfv{x})$ that represent two ``knobs'' one can
tune to improve gradient accuracy. We hope our work may lead to a systematic
pursuit to characterizing all possible gradient estimators based on different
principles of Monte Carlo gradient estimation.

\subsubsection*{Acknowledgements}
PP was supported by KAKENHI Grants 16H06561, 16H06563 and 16K21738,
by the research support of the Okinawa Institute of Science and Technology
Graduate University for the Neural Computation Unit, by RIKEN
during an internship, and by the Cyborg-AI project supported by NEDO, Japan.
MS was supported by KAKENHI 17H00757.

\bibliography{unifgrad}
\bibliographystyle{apalike}


\onecolumn

\appendix
\renewcommand{\appendixpagename}{}
\appendixpage

\aistatstitle{A unified view of likelihood ratio and reparameterization
  gradients: \\
Supplementary Materials}

\startcontents[sections]
\printcontents[sections]{l}{1}{\setcounter{tocdepth}{2}}
\newpage
\section{Additional background literature on LR and RP}
\label{additionalbackground}

See the work by \citet{mcgradrev} for a recent extensive general
review of Monte Carlo gradient estimators, such as LR and RP. Here we
discuss the main points about LR and RP in the literature, and how
these are related to our work.
\paragraph{Advantages and disadvantages of LR and RP:}
The variance of LR and RP gradients has been of central importance in
their research. Typically, RP is said to be more accurate and scale
better with the sampling dimension
\citep{rezende2014stochasticBP}---this claim is also backed by theory
\citep{xu2018rpvar,nesterov2017randomtheory}; however, {\bf there is no
  guarantee that RP outperforms LR}. In particular, for multimodal
$\phi(x)$ \citep{gal2016uncertainty} or chaotic systems \citep{pipps},
LR can be arbitrarily better than RP (e.g., the latter showed that LR
can be $10^6$ more accurate in practice). Moreover, RP is not directly
applicable to discrete sampling spaces, but requires continuous
relaxations
\citep{maddison2016concrete,jang2016categorical,tucker2017rebar}. Differentiable
RP is also not always possible, but implicit RP gradients have
increased the number of usable distributions
\citep{figurnov2018implicitRP}. Because LR and RP both have advantages
and disadvantages, optimal estimation techniques will require a combination
of LR and RP as considered in the flow gradients.
\remove{they are not directly comparable, and improvements
to them should be considered orthogonally, which is why we did not
compare to RP gradients in the main text. Clearly, in the example with
a quadratic $\phi(\bfv{x})$, RP would scale much better with the
dimension; however, in situations such as the evolution strategies
experiments in App~\ref{evolstrategies}, RP is not applicable because
$\nabla_{\bfv{x}}\phi(\bfv{x})$ is not available. Moreover, it is easy
to construct situations where the RP gradient variance explodes, but
LR stays accurate, by changing $\phi(x)$ to
$\phi(x) + a\sin(\omega x)$, where $a$ is small, but $\omega$ is
large.}
\paragraph{Variance reduction techniques:}
Techniques for variance reduction have
been extensively studied, including control variates/baselines
\citep{grathwohl2017bpthroughvoid,greensmith2004cv,tucker2018mirage,gu2015muprop,geffner2018using,gu2016q}
as well as Rao-Blackwellization 
\citep{aueb2015local,ciosek2018expected,asadi2017meanac}. One can also
combine the best of both LR and RP gradients by dynamically
reweighting them \citep{pipps,metz2019und}.

\paragraph{LR and RP on computational graphs:}
Several methods for
computing LR and RP gradients on graphs of computations exist
\citep{schulman2015stocgraph,weber2019credit,parmas2018total,foerster2018dice,
  mao2019baseline,farquhar2019loaded}. 
Among these works, \citet{schulman2015stocgraph} provided a simple way
to obtain the gradient estimators using automatic differentiation of a
surrogate objective on stochastic computation graphs;
\citet{weber2019credit} extended this work to be also applicable for
gradient estimators using critics; \citet{parmas2018total} provided an
intuitive abstract framework for reasoning about gradient estimators
by turning the stochastic graph deterministic through considering
gradients of the marginal distributions w.r.t.\ the distributions at
the other nodes, thus allowing to apply the total derivative rule;
\citet{foerster2018dice} extended the surrogate loss concept for
higher order derivatives; \citet{mao2019baseline} provided a baseline
for higher order LR estimators, and \citet{farquhar2019loaded} derived
a way to trade off bias and variance.
\paragraph{Importance sampling:}
Importance sampling for reducing LR gradient variance was previously
considered in variational inference \citep{ruiz2016impinvarinf}, who
 proposed to sample from the same distribution while tuning the
variance. In
reinforcement learning, importance sampling has been studied for
sample reuse via off-policy policy evaluation
\citep{thomas2016offpolicy,jiang2016doubly,gu2017interpolatedpol,munos2016safe,jie2010connection},
but modifying the policy to improve gradient accuracy has not been
considered.

\paragraph{Other:}
The flow theory in Sec.~\ref{flowtheor} was concurrently derived by
\citep{jankowiak2018rpflow}, but their work focused on deriving new RP
gradient estimators, and they do not discuss the duality. Our
derivation is also more visual. We discuss the relationship between
their work and ours in more detail in App.~\ref{flowgradproofs}.  The
flow gradient estimator in Eq.~(\ref{probflowgrad}) was also presented
in a work in progress by \citet{wu2019generalized} slightly earlier,
but in their derivation they do not use the divergence theorem, and
assume that $\bfv{u}_{\theta_i}(\bfv{x})$ belongs to a
reparameterizable variable, whereas we do not make such an assumption,
and argue that the flow is a more fundamental concept than the
reparameterization of the variable.  An advantage of their work is
that they show a new polynomial-based gradient estimator, whereas we
focus on characterizing the space of estimators.  One point is that
there always exists a flow corresponding to a reparameterization;
however, it is an open question whether there exists a
reparameterization corresponding to each flow
$\bfv{u}_{\theta_i}(\bfv{x})$. Our main contribution regarding this
estimator is the discussion around it, which claims that the flow
$\bfv{u}_{\theta_i}(\bfv{x})$ is a fundamental concept that
characterizes the space of all possible single sample estimators, as
well as the visualizations and physical intuitions that we provide.

\section{Likelihood ratio gradient basics}
\label{lrbasicsapp}

The likelihood ratio (LR) gradient estimator is given by

\begin{equation}
  \deriv{}{\theta}\expectw{x\sim\p{x;\theta}}{\phi(x)} =
  \expectw{x\sim\p{x;\theta}}{\deriv{\log\p{x;\theta}}{\theta}\phi(x)}.
\end{equation}

For a Gaussian $\p{x;\theta}$:

\begin{equation}
\begin{aligned}
  \log\p{x;\theta} &= -\frac{1}{2}\log(2\pi) - \log(\sigma) -
  \frac{(x-\mu)^2}{2\sigma^2},\\
  \deriv{\log\p{x;\theta}}{\mu} &= \frac{x-\mu}{\sigma^2} =
  \frac{\epsilon}{\sigma},\\
  \deriv{\log\p{x;\theta}}{\sigma} &= \frac{(x-\mu)^2}{\sigma^3} -
  \frac{1}{\sigma} = \frac{\epsilon^2}{\sigma} - \frac{1}{\sigma},\\
  \textup{where }~~&x=\mu + \epsilon\sigma \textup{ and }\epsilon\sim\mathcal{N}(0,1).
\end{aligned}
\end{equation}

\paragraph{Baselines to reduce gradient variance:} The LR gradient
estimator on its own has a large variance, and techniques have to be used
to stabilize it. A common technique is to subtract a constant baseline $b$
from the $\phi(x)$ values, so that the gradient estimator becomes

\begin{equation}
  \deriv{}{\theta}\expectw{x\sim\p{x;\theta}}
  {\deriv{\log\p{x;\theta}}{\theta}\left(\phi(x)-b\right)}.
\end{equation}

In practice, using $b = \expectw{x\sim\p{x;\theta}}{\phi(x)}$ works
well, but one can also derive an optimal baseline
\citep{weaver2001optimalbaseline}. We outline the derivation below. The
gradient variance when a baseline is used can be expressed as

\begin{equation}
\label{eq:gradvarderiv}
  \begin{aligned}
    \variancew{x\sim\p{x;\theta}}
    {\deriv{\log\p{x;\theta}}{\theta}\left(\phi(x)-b\right)} &=
    ~~\expectw{x\sim\p{x;\theta}}{\left(\deriv{\log\p{x;\theta}}{\theta}\phi(x)\right)^2}\\
    &-2\expectw{x\sim\p{x;\theta}}{\left(\deriv{\log\p{x;\theta}}{\theta}\right)^2\phi(x)b}\\
    &+
    \expectw{x\sim\p{x;\theta}}{\left(\deriv{\log\p{x;\theta}}{\theta}b\right)^2}.
  \end{aligned}
\end{equation}

Taking the derivative of Eq.~(\ref{eq:gradvarderiv}) w.r.t.\ $b$ and setting
to zero gives the optimal baseline as

\begin{equation}
  b_{opt} = \frac{
    \expectw{ x\sim\p{x;\theta} }
    { \left(\deriv{\log\p{x;\theta}}{\theta}
      \right)^2\phi(x) }
  }
  {\expectw{ x\sim\p{x;\theta} }
    { \left(\deriv{\log\p{x;\theta}}{\theta}
      \right)^2}}.
\end{equation}

In practice, for example if $\phi(x)$ is linear and $\p{x;\theta}$ is
 Gaussian then $b_{opt} = \expectw{x\sim\p{x;\theta}}{\phi(x)}$, so
the gain from trying to use an optimal baseline is often small.  

\paragraph{Antithetic sampling:} An often used technique is to sample
points $x$ in pairs opposite to each other, s.t.\
$x_+ = \mu + \sigma\epsilon$ and $x_- = \mu - \sigma\epsilon$. This
technique is particularly often used in evolution strategies' research
\citep{salimans2017oaies,mania2018simplers}. We will explain that when
this technique is used, then a baseline has no effect because it cancels.
The derivation is easy to see by
considering that for a Gaussian:
$\deriv{\log\p{x;\theta}}{\mu} = \frac{\epsilon}{\sigma}$, so
\[\begin{aligned}\deriv{\log\p{x_+;\theta}}{\mu}(\phi(x_+)-b) +
\deriv{\log\p{x_-;\theta}}{\mu}(\phi(x_-)-b) \\=
\frac{\epsilon}{\sigma}\left(\phi(x_+) -b - (\phi(x_-) - b)\right) =
\frac{\epsilon}{\sigma}\left(\phi(x_+) - \phi(x_-)\right).
\end{aligned}\]

\paragraph{Relationship to finite difference methods:} Finite
difference methods also use the function values $\phi(x)$ to estimate
a derivative, so it may appear that the LR gradient estimator is a
finite difference estimator. Finite difference estimators work by
estimating the slope of the function, by evaluating the change between
two points, i.e.\
\begin{equation}
\deriv{\phi(x)}{x} \approx \frac{\phi(x_+) - \phi(x_-)}{\Delta x}.
\end{equation}
In the antithetic sampling case, $\Delta x = 2\sigma\epsilon$, so
the estimator is 
\begin{equation}
\deriv{\phi(x)}{x} \approx \frac{\phi(x_+) - \phi(x_-)}{2\sigma\epsilon}.
\end{equation}
Clearly, this is different to the LR gradient estimator averaged over one
pair of samples
$\frac{1}{2}\left(\frac{\epsilon}{\sigma}\phi(x_+) +
  \frac{-\epsilon}{\sigma}\phi(x_-)
\right)$:
\begin{equation}
  \frac{\epsilon\left(\phi(x_+) - \phi(x_-)\right)}{2\sigma},
\end{equation}
because the $\epsilon$ is in the wrong place. In Sec.~\ref{boxtheor} we explain
that the LR gradient estimator is a different concept to finite differences, which
is not trying to fit a linear function onto $\phi(x)$.

\section{Probability boxes formal derivation}
\label{probboxapp}

Here we give a more formal derivation of the ``probability boxes''
explanation in Sec.~\ref{boxtheor}. In the theory, we explained that
the expectation can be written as a weighted average $\sum_{i=1}^NP(x_i)\phi(x_i)$
of function values, where the weight is given by the probabilities. Then,
the LR gradient estimated the derivative of this expectation w.r.t.\ $\theta$
by differentiating the $P(x_i)$ term, whereas the RP gradient worked by
differentiating the $\phi(x_i)$ term. In this section, we will formally
define the locations and edges of the ``boxes'' when a continuous integral
is discretized, and show that in the
LR case, indeed $\phi(x_i)$ in the ``box'' stays fixed, whereas in the RP case,
indeed the probability mass $P(x_i)$ in the ``box'' stays fixed as the
parameters $\theta$ are perturbed.

The explanation relies on a first principles
thinking about the effect that changing the parameters of a
probability distribution $\theta$ has on infinitesimal ``boxes" of
probability mass (Fig.~\ref{probboxes}). 
Both LR and RP are trying to estimate
\begin{equation}
  \deriv{}{\theta}\int \p{x;\theta}\phi(x)\textup{d}x.
\end{equation}
A typical
finite explanation of Riemann integrals is performed by discretizing
the integrand into ``boxes" of size $\Delta x$, and summing:
\begin{equation}
  \deriv{}{\theta}\sum_{i=1}^N \p{x_i;\theta}\Delta x_i\phi(x_i).
\end{equation}
Taking the limit as $N \rightarrow \infty$ recovers the true integral.
In this equation, $P(x_i) = \p{x_i;\theta}\Delta x_i$ is the amount of
probability mass inside the ``box", and $\phi(x_i)$ is the function
value inside the ``box".

\paragraph{RP estimator:}
Such a view can be used to explain RP gradients.  In this case, the
boundaries of the ``box" are fixed with reference to the shape of the
probability distribution, i.e.\ for each $i$ we define the center of
the box as
\begin{equation}
  x_i = g(\epsilon_i;\theta),
\end{equation}
and the boundaries as
\begin{equation}
  g(\epsilon_i\pm \Delta\epsilon/2;\theta),
\end{equation}
where $\epsilon_i$ is the
reference position on a fixed simple distribution, $\p{\epsilon}$.  The
amount of probability mass assigned to each ``box" stays fixed at
\begin{equation}
  P_i = \p{\epsilon}\Delta\epsilon~;
\end{equation}
however, the center of the
``box" moves, so the function value $\phi(x_i)$ inside each ``box"
changes by
\begin{equation}
  \delta\phi_i =
\phi\left(g(\epsilon_i;\theta+\delta\theta)\right)~-~\phi\left(g(\epsilon_i;\theta)\right)
= \phi(x_i~+~\delta x_i)~-~\phi(x_i).
\end{equation}
The full derivative can then be
expressed as
\begin{equation}
  \deriv{}{\theta}\expectw{x\sim\p{x;\theta}}{\phi(x)} \approx
\frac{1}{\delta\theta}\sum_{i=1}^NP_i\delta\phi_i =
\sum_{i=1}^NP_i\frac{\delta\phi_i}{\delta x_i}\frac{\delta
  x_i}{\delta\theta}.
\end{equation}
Taking the infinitesimal limit
$N\rightarrow\infty$, and noting
$P_i = \p{x_i;\theta}\Delta x_i$, we obtain the RP estimator
\begin{equation}
  \int \p{x;\theta}\deriv{\phi(x)}{x}\deriv{x}{\theta}~\textup{d}x.
\end{equation}
We see that RP essentially estimates the gradient by keeping the
probability mass inside each ``box" fixed, but estimating how the
function value $\phi$ inside the ``box" changes as the parameters
$\theta$ are perturbed.

\remove{
As the parameters of the probability
distribution are shifted by $\delta\theta$, the RP method keeps the
amount of probability mass inside each ``box" fixed, but the position
of the ``box" changes meaning that the function value $\phi(x_i)$
inside each ``box" changes. The gradient is estimated by estimating
this change $\deriv{\phi(x_i)}{x}\delta x_i$.
}

\paragraph{LR estimator:} The LR gradient, on the other hand, keeps the boundaries of the
``boxes" fixed, i.e.\ the center of the box is at $x_i$, and the
boundaries at \begin{equation}
  x_i \pm \Delta x_i/2.
\end{equation}
Now, as the boundaries are
independent of $\theta$, the function value $\phi(x_i)$ inside the box
stays fixed, even as $\theta$ is perturbed by $\delta\theta$; however,
the probability mass inside the box changes, because the density
changes by
\begin{equation}
  \delta p_i = \p{x_i;\theta+\delta\theta} - \p{x_i;\theta}.
\end{equation}
The full
derivative can be expressed as
\begin{equation}
  \deriv{}{\theta}\expectw{x\sim\p{x;\theta}}{\phi(x)} \approx
  \frac{1}{\delta\theta}\sum_{i=1}^N\Delta x_i\delta p_i\phi(x_i) =
\sum_{i=1}^N \p{x_i;\theta}\Delta x_i\frac{\delta
  p_i/\delta\theta}{\p{x_i;\theta}}\phi(x_i).
\end{equation}
Where we have multiplied and divided by $\p{x_i;\theta}$. Taking the
infinitesimal limit recovers the LR gradient
\begin{equation}
  \int\p{x;\theta}\frac{\deriv{\p{x;\theta}}{\theta}}
{\p{x;\theta}}\phi(x)~\textup{d}x =
\expectw{x\sim\p{x;\theta}}{\frac{\deriv{\p{x;\theta}}{\theta}}
  {\p{x;\theta}}\phi(x)}.
\end{equation}
The transformation
\begin{equation}
  \p{x;\theta}\frac{\deriv{\p{x;\theta}}{\theta}} {\p{x;\theta}} =
  \p{x;\theta}\deriv{\log\p{x;\theta}}{\theta}
\end{equation}
is known as the
log-derivative trick, and it may appear to be the essence behind the
LR gradient, but actually the multiplication and division by
$p(x;\theta)$ is just a special case of the more general Monte Carlo
integration principle. Any integral $\int f(x)~\textup{d}x$ can be
approximated by sampling from a distribution $q(x)$ as
\begin{equation}
  \int f(x)~\textup{d}x = \int q(x) \frac{f(x)}{q(x)}~\textup{d}x =
  \expectw{x\sim q(x)}{\frac{f(x)}{q(x)}}.
\end{equation}
Rather than thinking of the
LR gradient in terms of the log-derivative term, it may be better
to think of it as simply estimating the integral
$\int\deriv{\p{x;\theta}}{\theta}\phi(x)~\textup{d}x$ by applying the
appropriate importance weights to samples from $p(x;\theta)$. Thus, we
see that in the discretized case, the LR gradient picks
$q(x) = \p{x;\theta}$ \citep{jie2010connection} and performs Monte
Carlo integration to approximate
$\frac{1}{\delta\theta}\sum_{i=1}^N\Delta x_i\delta p_i\phi(x_i)$ by
sampling from $P(x_i) = \Delta x_i\p{x_i;\theta}$. To summarize: LR
estimates the gradient by keeping the boundaries of the boxes fixed,
measuring the change in probability mass in each box, and weighting by
the function value: $\phi(x_i)\delta p$.

Sometimes, the LR gradient is described as being ``kind of like a
finite difference gradient"
\citep{salimans2017oaies,mania2018simplers}, but here we see that it is
a different concept, which does not rely on fitting a straight line
between differences of $\phi$ (App.~\ref{lrbasicsapp}), but estimates
how probability mass is reallocated among different $\phi$ values via
Monte Carlo integration by sampling from $\p{x;\theta}$.

\section{Derivations for the probability flow theory}
\label{probflowderivapp}
(The notation and divergence theorem proof are the same as in the main
paper, and provided here for reference.) We illustrate the
background information in 3 dimensions, but it generalizes
straightforwardly to higher dimensions.

\paragraph{Notation:}\mbox{}\\
$\bfv{F} = [F_x(x,y,z), F_y(x,y,z), F_z(x,y,z)]$ is a vector field. \\
$\phi(x,y,z)$ is a scalar field (a scalar function).\\
Div operator:
$\nabla\cdot \bfv{F} = \pderivw{F_x}{x} + \pderivw{F_y}{y} +
\pderivw{F_z}{z}$. \\Grad operator:
$\nabla\phi = [\pderivw{\phi}{x}, \pderivw{\phi}{y},
\pderivw{\phi}{z}]$.

\subsection{Basic vector calculus and fluid mechanics}

\begin{figure*}[!t]
        \centering
\includegraphics[width=.6\textwidth]{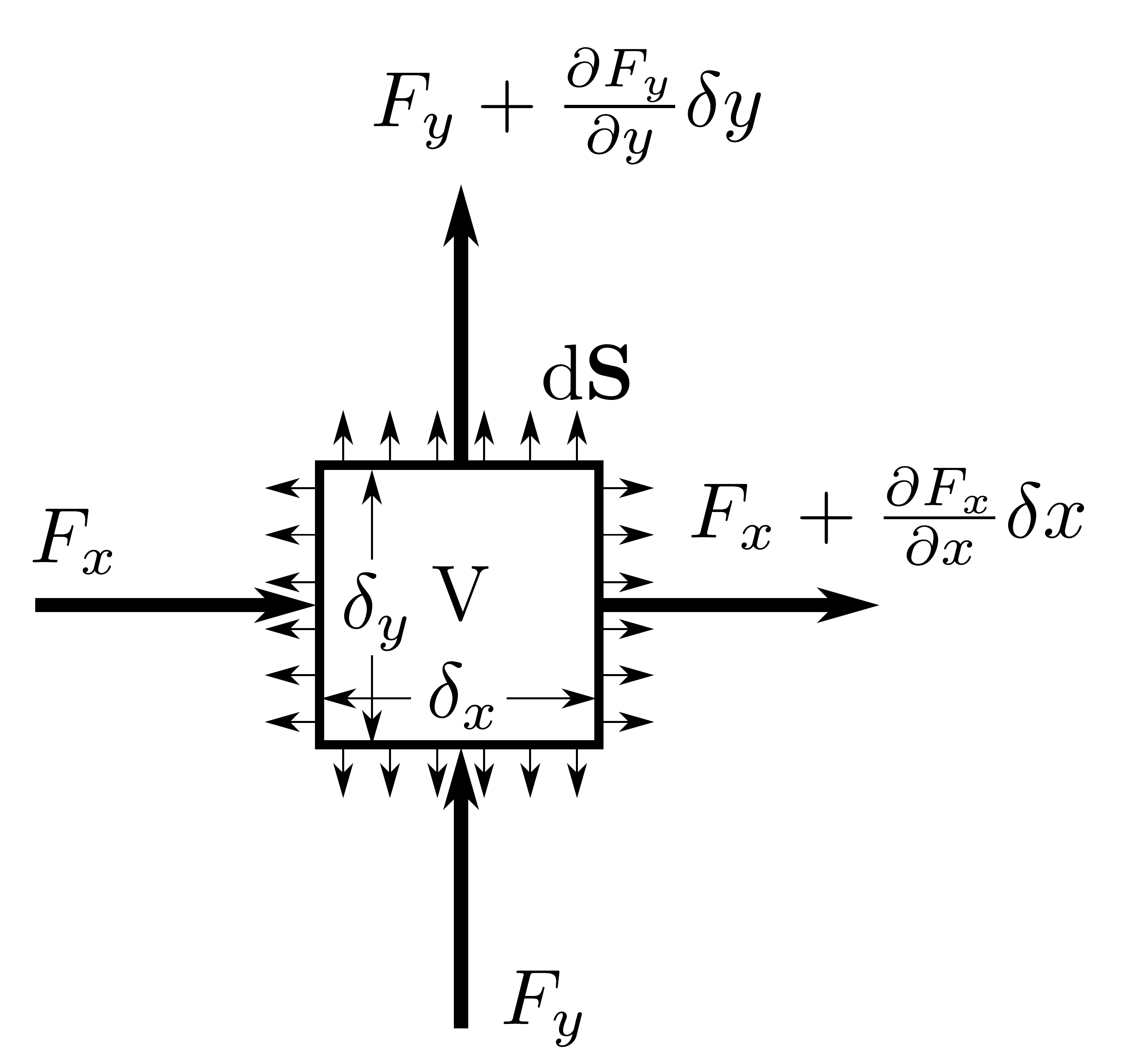}
\caption{Illustration of the divergence theorem.}
          \label{divtheorem}
\end{figure*}

The vector field $\bfv{F}$ could be for example thought of as a local
flow velocity for some fluid. If $\bfv{F}$ is the density flow rate, then
the div operator essentially measures how much the density is decreasing at
a point. If the outflow is larger than the inflow, the density would
decrease and vice versa. The divergence theorem, illustrated in Fig.~\ref{divtheorem}
illustrates how this change in density can be measured in two separate ways:
one could integrate the divergence across the volume, or one could integrate the
inflow and outflow across the surface. The divergence theorem states:

\begin{equation}
  \int_V \nabla\cdot\bfv{F}\textup{d}V = \int_S\bfv{F}\cdot\textup{d}\bfv{S}.
\end{equation}

To prove the claim, consider the infinitesimal box in
Fig.~\ref{divtheorem}.  The divergence can be calculated as
$\delta x\delta y(\pderivw{F_x}{x} + \pderivw{F_y}{y})$. On the other
hand, to take the integral across the surface, note that the surface
normals point outwards, and the integral becomes
$\delta_y(-F_x + F_x - \pderivw{F_x}{x}\delta x) + \delta x(-F_y + F_y
+ \pderivw{F_y}{y}\delta y) = \delta x\delta y(\pderivw{F_x}{x} +
\pderivw{F_y}{y})$, which is the same as the divergence. To generalize this
to arbitrarily large volumes, notice that if one stacks the boxes next to each
other, then the surface integral across the area where the boxes meet cancels
out, and only the integral across the outer surface remains. For an
incompressible flow, the density does not change, and the divergence
must be zero.

\subsection{Derivation of probability surface integral}
  \label{surfDeriv}
  We will show that the LR estimator tries to integrate
  $\int_S\phi( \tilde{\bfv{x}})\nabla_\theta
  \tilde{g}(\epsilon_x,\epsilon_h)\textup{d}\bfv{S}$. First, note that
  $\textup{d}\bfv{S} = \hat{\bfv{n}}\textup{d}S$, and it is necessary
  to express the normalized surface vector $\hat{\bfv{n}}$. To do so,
  we first express the tangent vectors $\bfv{t}$, then find the vector
  perpendicular to all of them (this is exactly the normal vector).
  \remove{ A vector tangent and downhill to the surface is given by
    $\bfv{t} = [-\dpx,-\left(\dpx\right)\left(\dpx\right)^T]$. The
    normal vector $\bfv{n}$ is $[-\dpx;h]$, such that
    $\bfv{t}\cdot\bfv{n} = 0$.  Therefore,
    $\left(\dpx\right)\left(\dpx\right)^T -
    \left(\dpx\right)\left(\dpx\right)^Th = 0 ~~\Rightarrow~~ h = 1$.
    Finally, we normalize the vector:
\begin{equation}
  \hat{\bfv{n}} = \nicefrac{[-\dpx,1]}{\sqrt{\left(\dpx\right)
      \left(\dpx\right)^T+ 1}}~.
  \label{eq:n}
\end{equation}
}

All tangent vectors are characterized by the equation
$\bfv{t} = \left[\bfv{r},\dpx\cdot\bfv{r}\right]$, where $\bfv{r}$ is
an arbitrary vector. The normal vector $\bfv{n}$ is such that
$\bfv{t}\cdot\bfv{n} = 0$ for all tangent vectors. Therefore, the
normal vector $\bfv{n} = \left[-\dpx,1\right]$, satisfies the equation, because
$\bfv{n}\cdot\bfv{t} = -\dpx\cdot\bfv{r} + 1\cdot\dpx\cdot\bfv{r} =
0$.  Finally, we normalize the vector:
\begin{equation}
  \hat{\bfv{n}} = \nicefrac{\left[-\dpx,1\right]}{\sqrt{\left(\dpx\right)
      \left(\dpx\right)^T+ 1}}~.
  \label{eq:n}
\end{equation}

Next, we perform a change of coordinates from the surface elements
$\textup{d}S$ to cartesian coordinates $\textup{d}\bfv{x}$.  When
projecting a surface element $\textup{d}S$ with unit normal
$\hat{\bfv{n}}$ to a plane with unit normal $\hat{\bfv{m}}$, the
projected area is given by
$\textup{d}\bfv{x} =
\left|\hat{\bfv{n}}\cdot\hat{\bfv{m}}\right|\textup{d}S$, therefore,
as $\hat{\bfv{m}} = [\bfv{0},1]$ for the $\bfv{x}$-plane, we have
$\textup{d}\bfv{x}~=~\textup{d}S\left|\frac{1}{\sqrt{\left(\dpx\right)
      \left(\dpx\right)^T+ 1}}\left[-\dpx,1\right]\cdot[\bfv{0},1]\right| =
\nicefrac{\textup{d}S}{\sqrt{\left(\dpx\right) \left(\dpx\right)^T+ 1}}
$, which leads to
\begin{equation}
  \textup{d}S =
  \sqrt{\left(\dpx\right) \left(\dpx\right)^T+ 1}~~\textup{d}\bfv{x}.
  \label{eq:dS}
\end{equation}

Plugging Eqs.~(\ref{eq:n}) and (\ref{eq:dS}) into the right-hand side
of Eq.~(\ref{surfint}) we get
\begin{equation}
\begin{aligned}
\int_{X} \phi(
\tilde{\bfv{x}})\nabla_\theta \tilde{g}(\epsilon_x,\epsilon_h)
\cdot
\frac{\left[-\dpx,1\right]}{\sqrt{\left(\dpx\right)
      \left(\dpx\right)^T+ 1}}
  \sqrt{\left(\dpx\right) \left(\dpx\right)^T+ 1}~~\textup{d}\bfv{x}
  =
\int_{X} \phi(
\tilde{\bfv{x}})\nabla_\theta \tilde{g}(\epsilon_x,\epsilon_h)
\cdot
\left[-\dpx,1\right]~\textup{d}\bfv{x}.
\label{eq:normprod}
\end{aligned}
\end{equation}

Recall that the last element of $\tilde{g}(\epsilon_x,\epsilon_h)$ is
$\epsilon_h\p{g(\epsilon_x);\theta}$, and that at the
boundary surface $\epsilon_h = 1$, then the scalar product term
$\nabla_\theta \tilde{g}(\epsilon_x,\epsilon_h)\cdot \left[-\dpx,1\right]$ turns
into
$-\nabla_\theta g(\epsilon_x)\cdot\dpx +
\pderiv{\epsilon_h\p{g(\epsilon_x);\theta}}
{\theta}{\epsilon_x=const,\epsilon_h=1}$. The last term
$\pderiv{\p{g(\epsilon_x);\theta}}
{\theta}{\epsilon_x=const}$ can be thought of as the rate
of change of the probability density while following a point moving in the flow
induced by perturbing $\theta$. This quantity can be expressed with
the material derivative
$\pderiv{\p{g(\epsilon_x);\theta}}
{\theta}{\epsilon_x=const} =
\deriv{\p{\bfv{x};\theta}}{\theta} + \nabla_\theta
g(\epsilon_x)\cdot\dpx$. Finally, substituting into Eq.~(\ref{eq:normprod}):

\begin{equation}
\begin{aligned}
  \int_{S} \phi(\tilde{\bfv{x}})\nabla_\theta
  \tilde{g}(\epsilon_x,\epsilon_h)~\textup{d}\bfv{S} =
  \int_{X}
  \phi(\bfv{x})\deriv{\p{\bfv{x};\theta}}{\theta}
  ~\textup{d}\bfv{x}.
\label{eq:subsapp}
\end{aligned}
\end{equation}

\remove{We have already seen that a Monte Carlo integration of
the right hand side of Eq.~(\ref{eq:subs}) using samples from
$\p{\bfv{x};\theta}$ gives rise to the LR gradient
estimator. Thus, the RP gradient estimator and the LR gradient
estimator are duals under the divergence theorem. To further strengthen this
claim we prove that the LR gradient estimator is the unique estimator
that takes weighted averages of the function values $\phi(\bfv{x})$.}

\section{Characterizing the space of all likelihood ratio and reparameterization gradients}

\subsection{Reparameterization gradients are not unique}

\label{rpNotUnique}

What happens if we perform the same kind of analysis as in
Theorem~\ref{lrunique} for the RP gradient? To examine this idea, first we
extend the argumentation in the main paper to include non-invertible $g(\epsilon;\theta)$,
i.e., consider the case when multiple different $\epsilon$ lead to the same $x$ via $x = g(\epsilon;\theta)$.
In Sec.~\ref{coordinatetrans} we argued that if $g$ is invertible, we can write the
RP gradient as $\deriv{\phi(x)}{x}\pderiv{g(\epsilon;\theta)}{\theta}{\epsilon=S(x;\theta)}$,
where $\pderiv{g(\epsilon;\theta)}{\theta_i}{\epsilon=S(x;\theta)}$ corresponds
to $\bfv{u}_{\theta_i}(\bfv{x})$, which is a vector field. In the non-invertible case,
similarly denote $S(x;\theta)$ is the set of $\epsilon$, s.t.\ $x = g(\epsilon;\theta)$.
For each x, we can employ Bayes' rule to derive the posterior distribution of the $\epsilon$
that generated $x$, and integrate across this distribution to obtain the Rao-Blackwellized
estimator
$\deriv{\phi(x)}{x}\left.\int_{S(x;\theta)}p(\epsilon)\pderiv{g(\epsilon;\theta)}{\theta}{\epsilon\in S(x;\theta)}\textup{d}\epsilon
\middle/\int_{S(x;\theta)}p(\epsilon)\textup{d}\epsilon\right.$, where
$\left.\int_{S(x;\theta)}p(\epsilon)\pderiv{g(\epsilon;\theta)}{\theta_i}{\epsilon\in S(x;\theta)}\textup{d}\epsilon
\middle/\int_{S(x;\theta)}p(\epsilon)\textup{d}\epsilon\right.$ corresponds to
$\bfv{u}_{\theta_i}(\bfv{x})$, which is a vector field. Thus, even in the non-invertible case, the
reparameterization gradient can be expressed as a dot product between $\nabla_{\bfv{x}}\phi(\bfv{x})$ and
a vector field $\bfv{u}(\bfv{x})$.

Now we show that the type of analysis in Thm.~\ref{lrunique} does not lead to a uniqueness claim
for RP gradients. Similarly, suppose that
there exist $\bfv{u}(\bfv{x})$ and $\bfv{v}(\bfv{x})$, s.t.\
$\int \nabla\phi(\bfv{x})\cdot \bfv{u}(\bfv{x})~\textup{d}\bfv{x} = \int
\nabla\phi(\bfv{x})\cdot \bfv{v}(\bfv{x})~\textup{d}\bfv{x}$ for any
$\phi(\bfv{x})$.  Rearrange the equation into
$\int
\nabla\phi(\bfv{x})\cdot\left(\bfv{u}(\bfv{x})-\bfv{v}(\bfv{x})\right)~\textup{d}\bfv{x}
= 0$.  Then, if we can pick
$\nabla\phi(\bfv{x}) = \bfv{u}(\bfv{x})-\bfv{v}(\bfv{x})$ it would lead to $\bfv{u}=\bfv{v}$,
which would show the uniqueness. In the one-dimensional case, this is possible, and in this case all
pure reparameterization or pathwise gradients are equivalent. However,
in higher dimensions, it is not necessarily
possible to pick such $\phi(\bfv{x})$. In particular, the integral of
$\nabla\phi(\bfv{x})$ over any closed path is 0, but this is not
necessarily the case for $\bfv{u}-\bfv{v}$. Therefore, the same kind of analysis
does not lead to a claim of uniqueness. Indeed, concurrent work
\citep{jankowiak2018rpflow} showed that there are an infinite amount
of possible reparameterization gradients, and the minimum
variance\footnote{By minimum variance, we mean the minimum variance
  achievable without assuming knowledge of $\phi(\bfv{x})$, or
  alternatively that it is approximately linear in the sampling range,
  $\nabla\phi(\bfv{x}) \approx \bfv{A}$. Their result holds for
  arbitrary dimensionality.} is achieved by the optimal transport
flow.

\subsection{Flow gradient estimator}
\label{flowgradproofs}
In this section, we provide the derivations and proofs for the flow
gradient estimator in Eq.~(\ref{probflowgrad}). While our derivation
with a height reparameterization in Sec.~\ref{flowtheor} shows the
duality between LR and RP, and allows for intuitive visualizations,
the method of derivations by \citet{jankowiak2018rpflow} is
algebraically easier to work with, so we adopt their style in the
following sections. Recall the divergence theorem
$ \int_V\nabla_{\bfv{x}}\cdot \bfv{F}\textup{d}V = \int_S \bfv{F}\cdot
\textup{d}\bfv{S},$ where $\bfv{F}$ is an arbitrary continuous
piecewise differentiable vector field. We can pick
$\bfv{F} =
\p{\bfv{x};\theta}\bfv{u}_{\theta_i}(\bfv{x})\phi(\bfv{x})$, and a surface $S$
at infinity, enclosing the whole volume, which
leads to the equation
\begin{equation}
\label{appdivF}
  \int_V\nabla_{\bfv{x}}\cdot\left(\p{\bfv{x};\theta}\bfv{u}_{\theta_i}(\bfv{x})\phi(\bfv{x})\right)\textup{d}V = \int_S\p{\bfv{x};\theta}\phi(\bfv{x})\bfv{u}_{\theta_i}(\bfv{x})\cdot
  \textup{d}\bfv{S}.
\end{equation}

Note that the boundary $S$ is at
$\lVert\bfv{x}\rVert \rightarrow \infty$, and in this case
$\p{\bfv{x};\theta} \rightarrow 0$, meaning that
$\int_S\p{\bfv{x};\theta}\phi(\bfv{x})\bfv{u}_{\theta_i}(\bfv{x})\cdot
\textup{d}\bfv{S} = 0$, as long as
$\phi(\bfv{x})\bfv{u}_{\theta_i}(\bfv{x})$ does not go to infinity
faster than $\p{\bfv{x};\theta}$ goes to $0$.\footnote{ Note that the
  case when
  $\p{\bfv{x};\theta}\phi(\bfv{x})\bfv{u}_{\theta_i}(\bfv{x})\not\to
  0$, does not correspond to any sensible estimator, because the value
  of $\phi(\bfv{x})$ at $\lVert\bfv{x}\rVert\to\infty$ will have an
  influence on the value of the gradient estimation. In that case,
  because $\p{\bfv{x};\theta}\to 0$ the probability of sampling at
  infinity will tend to $0$, and the gradient variance will explode. This condition does however mean that if one wants to construct
  a sensible estimator, care must be taken to ensure that
  $\bfv{u}_{\theta_i}(\bfv{x})$ does not go to infinity too fast, e.g. as
explained by \citet{jankowiak2019pathmulti}.}  Hence,
\begin{equation}
\label{flowgradeq}
\begin{aligned}
  \int_V\nabla_{\bfv{x}}\cdot\left(\p{\bfv{x};\theta}\bfv{u}_{\theta_i}(\bfv{x})\phi(\bfv{x})\right)\textup{d}V = 0\\
  \Rightarrow\int_V\biggl(\p{\bfv{x};\theta}\bfv{u}_{\theta_i}(\bfv{x})\cdot\nabla_{\bfv{x}}\phi(\bfv{x}) + \phi(\bfv{x})\nabla_{\bfv{x}}\cdot
  \bigl(\p{\bfv{x};\theta}\bfv{u}_{\theta_i}(\bfv{x})\bigr)\biggr)\textup{d}V = 0.
\end{aligned}
\end{equation}
As the integral in Eq.~(\ref{flowgradeq}) is 0, we can add it to
the integral of $\deriv{\p{\bfv{x};\theta}}{\theta_i}\phi(\bfv{x})$ without
changing the expectation:
\begin{equation}
\label{plusmintodtheta}
\begin{aligned}
  &\deriv{}{\theta_i}\expectw{\bfv{x}\sim\p{\bfv{x};\theta}}{\phi(\bfv{x})}\\
  &= 
  \int_X \biggl(\p{\bfv{x};\theta}\bfv{u}_{\theta_i}(\bfv{x})\cdot\nabla_{\bfv{x}}\phi(\bfv{x}) + \phi(\bfv{x})\nabla_{\bfv{x}}\cdot
  \bigl(\p{\bfv{x};\theta}\bfv{u}_{\theta_i}(\bfv{x})\bigr)\biggr)
  + \deriv{\p{\bfv{x};\theta}}{\theta_i}\phi(\bfv{x}) ~\textup{d}\bfv{x}\\
  &= \int_X \p{\bfv{x};\theta}\bfv{u}_{\theta_i}(\bfv{x})\cdot\nabla_{\bfv{x}}\phi(\bfv{x}) \textup{d}\bfv{x}
  + \int_X \biggl(\nabla_{\bfv{x}}\cdot
  \bigl(\p{\bfv{x};\theta}\bfv{u}_{\theta_i}(\bfv{x})\bigr)
  + \deriv{\p{\bfv{x};\theta}}{\theta_i}\biggr)\phi(\bfv{x})
  \textup{d}\bfv{x}.
\end{aligned}
\end{equation}
By importance sampling from $q(\bfv{x})$, the estimator is characterized
by the equation:
\begin{equation}
  \label{flowest}
  \begin{aligned}
  &\deriv{}{\theta_i}\expectw{\bfv{x}\sim\p{\bfv{x};\theta}}{\phi(\bfv{x})}
  \\
  &=
  \expectw{\bfv{x}\sim q(\bfv{x})}{
    \frac{\p{\bfv{x};\theta}}{q\left(\bfv{x}\right)}
    \bfv{u}_{\theta_i}(\bfv{x})\cdot\nabla_{\bfv{x}}\phi(\bfv{x})
    +
    \frac{1}{q\left(\bfv{x}\right)}\biggl(\nabla_{\bfv{x}}\cdot
  \bigl(\p{\bfv{x};\theta}\bfv{u}_{\theta_i}(\bfv{x})\bigr)
  + \deriv{\p{\bfv{x};\theta}}{\theta_i}\biggr)\phi(\bfv{x})
}.
\end{aligned}
\end{equation}
Next, we show that the estimator in Eq.~(\ref{flowest}) characterizes
the space of all possible single sample unbiased gradient estimators
that combine the function value $\phi(\bfv{x})$ and function
derivative $\nabla_{\bfv{x}}\phi(\bfv{x})$ information, and have the
product form given in Eq.~(\ref{esticlass}). The proof is
analogous to the proof of uniqueness of the LR gradient estimator in
Theorem~\ref{lrunique}. Remember that $\bfv{u}_{\theta_i}(\bfv{x})$ is
a completely arbitrary vector field (and hence covers all possible
ways to multiply with $\nabla_{\bfv{x}}\phi(\bfv{x})$ ); we show that
for each $\bfv{u}_{\theta_i}(\bfv{x})$ the corresponding weighting
function $\psi(\bfv{x})$ for $\phi(\bfv{x})$ that gives an unbiased
gradient estimator is unique, i.e.\
\[\psi(\bfv{x}) = \frac{1}{q\left(\bfv{x}\right)}\biggl(\nabla_{\bfv{x}}\cdot
\bigl(\p{\bfv{x};\theta}\bfv{u}_{\theta_i}(\bfv{x})\bigr) +
\deriv{\p{\bfv{x};\theta}}{\theta_i}\biggr).\]

\begin{theorem}[The flow gradient estimator characterizes the space
  of all single sample unbiased LR--RP estimators]
  Every unbiased gradient estimator of the form
  \[\deriv{}{\theta_i}\expectw{\bfv{x}\sim\p{\bfv{x};\theta}}{\phi(\bfv{x})}
    = \expectw{\bfv{x}\sim q(\bfv{x})}{
      \bfv{v}(\bfv{x})\cdot\nabla_{\bfv{x}}\phi(\bfv{x}) +
      \psi(\bfv{x})\phi(\bfv{x})},\] where $\phi(\bfv{x})$ is an
  arbitrary function, is a special case of the estimator characterized
  by
  \begin{equation*}\begin{aligned}
      \deriv{}{\theta_i}\expectw{\bfv{x}\sim\p{\bfv{x};\theta}}{\phi(\bfv{x})}
  =
  \mathbb{E}_{\bfv{x}\sim q(\bfv{x})}\biggl[
    &\frac{\p{\bfv{x};\theta}}{q\left(\bfv{x}\right)}
    \bfv{u}_{\theta_i}(\bfv{x})\cdot\nabla_{\bfv{x}}\phi(\bfv{x})
    \\
    &+ \frac{1}{q\left(\bfv{x}\right)}\biggl(\nabla_{\bfv{x}}\cdot
  \bigl(\p{\bfv{x};\theta}\bfv{u}_{\theta_i}(\bfv{x})\bigr)
  + \deriv{\p{\bfv{x};\theta}}{\theta_i}\biggr)\phi(\bfv{x})
\biggr].\end{aligned}
\end{equation*}
  \label{flowunique}
\end{theorem}
  \begin{proof}
    Note that there is a corresponding
    $\bfv{u}_{\theta_i}(\bfv{x})$ for an arbitrary $\bfv{v}(\bfv{x})$
    given by
    $\bfv{u}_{\theta_i}(\bfv{x})~=~\bfv{v}(\bfv{x})\frac{q\left(\bfv{x}\right)}{\p{\bfv{x};\theta}}$.
    We will show that $\psi(\bfv{x}) = \frac{1}{q\left(\bfv{x}\right)}\biggl(\nabla_{\bfv{x}}\cdot
  \bigl(\p{\bfv{x};\theta}\bfv{u}_{\theta_i}(\bfv{x})\bigr)
  + \deriv{\p{\bfv{x};\theta}}{\theta_i}\biggr)$
  is the unique function, s.t. the estimator with $\bfv{v}(\bfv{x})$ is
  unbiased.
    Suppose that there exist $\psi(\bfv{x})$ and $f(\bfv{x})$,
    s.t.\
    \[\int q(\bfv{x})\left(\bfv{v}(\bfv{x})\cdot\nabla_{\bfv{x}}\phi(\bfv{x}) +
      \psi(\bfv{x})\phi(\bfv{x})
      \right)~\textup{d}\bfv{x} =
      \int q(\bfv{x})\left(\bfv{v}(\bfv{x})\cdot\nabla_{\bfv{x}}\phi(\bfv{x}) +
      f(\bfv{x})\phi(\bfv{x})
      \right)~\textup{d}\bfv{x}\] for any $\phi(\bfv{x})$.
  Rearrange the equation into
  \[\int
  q(\bfv{x})\phi(\bfv{x})\left(\psi(\bfv{x})-f(\bfv{x})\right)~\textup{d}\bfv{x} =
  0,\] then pick $\phi(\bfv{x}) = \psi(\bfv{x})-f(\bfv{x})$ from which we
  get
  \[\int q(\bfv{x})\left(\psi(\bfv{x})-f(\bfv{x})\right)^2~\textup{d}\bfv{x} = 0.\]
  Therefore, $\psi = f$.
  As $\bfv{v}(\bfv{x})$ was arbitrary, and there is exactly one corresponding
  $\psi(\bfv{x})$ for each $\bfv{v}(\bfv{x})$, then the flow gradient
  estimator characterizes all possible single sample unbiased gradient
  estimators.
  \end{proof}

A corollary of Theorem~\ref{flowunique} is that the estimator is
independent of $\phi(\bfv{x})$ if and only if
\[\nabla_{\bfv{x}}\cdot\bigl(\p{\bfv{x};\theta}\bfv{u}_{\theta_i}(\bfv{x})\bigr) +
  \deriv{\p{\bfv{x};\theta}}{\theta_i} = 0,\]
which is the transport
equation required by the work of
\citet{jankowiak2018rpflow}. This means that the work by
\citet{jankowiak2018rpflow} characterized the space of all possible
RP gradients. On the other hand, if
$\bfv{u}_{\theta_i}(\bfv{x}) = \bfv{0}$, then we recover the unique LR gradient
estimator.

\subsection{A few examples of other estimators as
special cases of the flow gradient}
\label{flowgradexamples}
\paragraph{Notation:}\mbox{}\\
$g(\epsilon;\theta) = \bfv{x}$ is the reparameterization transformation.\\
$S(\bfv{x};\theta) = \epsilon$ is the standardization function (the inverse
of $g(\epsilon;\theta)$).\\
$\lvert X\rvert$ is the determinant of a matrix $X$.

\paragraph{Weighted average of LR and RP:}
Consider a gradient estimator given by
\begin{equation}
  \expectw{\bfv{x}\sim\p{x;\theta}}{k\nabla_{\bfv{x}}\phi(\bfv{x})
    \nabla_{\theta}g(\bfv{x}) + (1-k)\deriv{\log\p{x;\theta}}{\theta}\phi(\bfv{x})},
\end{equation}
where $k\in[0,1]$ is a weighting factor for the two gradient
estimators, and the gradient w.r.t.\ one parameter,
$\nabla_{\theta_i}g(\bfv{x})$, is defined as
$\pderiv{g(\epsilon;\theta)}{\theta_i}{\epsilon = S(\bfv{x};\theta)}$,
where $g(\epsilon;\theta)$ is a typical reparameterization. We will
show that this estimator belongs to the flow estimator class in
Eq.~(\ref{flowest}). The corresponding flow field for any $\theta_i$
is
$\bfv{u}_{\theta_i}(\bfv{x})~=~k\nabla_{\theta_i}g(\bfv{x})~=~k\pderiv{g(\epsilon;\theta)}{\theta}{\epsilon=S(\bfv{x};\theta)}$. Because
$g$ is a reparameterization, 
the flow without the $k$ factor,
$\tilde{\bfv{u}}_{\theta_i} = \nabla_{\theta_i}g(\bfv{x})$, satisfies
the transport equation (i.e.\ the $\phi(\bfv{x})$ term would disappear), 
$\nabla_{\bfv{x}}\cdot~\bigl(\p{\bfv{x};\theta}\tilde{\bfv{u}}_{\theta_i}(\bfv{x})
\bigr) + \deriv{\p{\bfv{x};\theta}}{\theta_i} = 0$, and hence,
$\nabla_{\bfv{x}}\cdot
\bigl(\p{\bfv{x};\theta}\bfv{u}_{\theta_i}(\bfv{x})\bigr) =
-k\deriv{\p{\bfv{x};\theta}}{\theta_i}$, which gives the desired
result when plugging into Eq.~(\ref{flowest}).

For clarity, we will show a specific example with a Gaussian
distribution, and considering the gradient w.r.t.\ one of the mean
parameters $\mu_i$, in which case $g = \mu + \sigma\epsilon$ and
$\nabla_{\mu_i}g(\bfv{x}) = [0,...,1,0,...,0]$, where the $1$ is at
the $i^{\textup{th}}$ dimension. In this case,
$\nabla_{\bfv{x}}\cdot\bigl(\p{\bfv{x};\theta}
\bfv{u}_{\theta_i}(\bfv{x})\bigr) =
\nabla_{\mu_i}g(\bfv{x})\cdot\nabla_{\bfv{x}}\p{\bfv{x};\theta} +
\p{\bfv{x};\theta}\nabla_{\bfv{x}}\cdot\nabla_{\mu_i}g(\bfv{x})$.
Note that $\nabla_{\bfv{x}}\cdot\nabla_{\mu_i}g(\bfv{x})=0$ because
$\nabla_{\mu_i}g(\bfv{x})$ is constant, and
$\nabla_{\mu_i}g(\bfv{x})\cdot\nabla_{\bfv{x}}\p{\bfv{x};\theta} =
\left[0,...,1,...,0\right]\cdot\left[\deriv{\p{\bfv{x};\theta}}{x_1},...,\deriv{\p{\bfv{x};\theta}}{x_i},
  ...,\deriv{\p{\bfv{x};\theta}}{x_D}\right] =
\deriv{\p{\bfv{x};\theta}}{x_i}$. Finally, note that
$\deriv{\p{\bfv{x};\theta}}{x_i} =
-\deriv{\p{\bfv{x};\theta}}{\mu_i}$, which gives the desired result.

\paragraph{Reparameterization gradient:} Here, we show explicitly
that the reparameterization flow,
$\bfv{u}_{\theta_i}(\bfv{x})~=~\pderiv{g(\epsilon;\theta)}{\theta}{\epsilon=S(\bfv{x};\theta)}$,
satisfies the transport equation, i.e.\
\begin{equation}
\label{RPflowcont}
  \nabla_{\bfv{x}}\cdot
  \left(\p{\bfv{x};\theta}\pderiv{g(\epsilon;\theta)}{\theta_i}{\epsilon=S(\bfv{x};\theta)}\right) +
  \deriv{\p{\bfv{x};\theta}}{\theta_i} = 0.
\end{equation}
First, we expand the divergence:
\begin{equation}
  \label{RPflownablaexp}
  \begin{aligned}
  \nabla_{\bfv{x}}\cdot
  \left(\p{\bfv{x};\theta}\pderiv{g(\epsilon;\theta)}{\theta_i}{\epsilon=S(\bfv{x};\theta)}\right)  =
  &\nabla_{\bfv{x}}\p{\bfv{x};\theta}\cdot\pderiv{g(\epsilon;\theta)}{\theta_i}{\epsilon=S(\bfv{x};\theta)} \\ &+
  \p{\bfv{x};\theta}\textup{Tr}\left[\pderivsq{g(\epsilon;\theta)}{\theta_i}
  {\bfv{\epsilon}}{\epsilon=S(\bfv{x};\theta)}\deriv{S(\bfv{x};\theta)}{\bfv{x}}\right].
\end{aligned}
\end{equation}
Consider the equation
\begin{equation}
\label{probepsilon}
  \p{\epsilon} = \p{\bfv{x};\theta}\left\lvert\deriv{g(\epsilon;\theta)}{\epsilon}\right\rvert = \p{g(\epsilon;\theta);\theta}\left\lvert\deriv{g(\epsilon;\theta)}{\epsilon}\right\rvert,
\end{equation}
where the determinant factor comes from the change of
coordinates. When differentiating Eq.~(\ref{probepsilon}) w.r.t.\
$\theta$, it becomes $0$, because $\p{\epsilon}$ does not depend on
$\theta$, and we will show that this gives rise to the transport
condition in Eq.~(\ref{RPflowcont}).
\begin{equation}
\begin{aligned}
\label{eq:pepszero}
  &\deriv{}{\theta_i}\left(\p{g(\epsilon;\theta);\theta}\left\lvert\deriv{g(\epsilon;\theta)}{\epsilon}\right\rvert\right) = 0\\
  &=\left(
    \pderiv{\p{g(\epsilon;\theta);\theta}}{\theta_i}{g=\bfv{x}}
    + \pderivw{\p{\bfv{x};\theta}}{\bfv{x}}\pderivw{g(\epsilon;\theta)}{\theta_i}\right)\left\lvert\deriv{g(\epsilon;\theta)}{\epsilon}\right\rvert
  + \p{g(\epsilon;\theta);\theta}
  \pderivw{}{\theta_i}\left(\left\lvert\deriv{g(\epsilon;\theta)}{\epsilon}\right\rvert\right)
\\
  &=\left(
    \pderivw{\p{\bfv{x};\theta}}{\theta_i}
    + \pderivw{\p{\bfv{x};\theta}}{\bfv{x}}\pderivw{g(\epsilon;\theta)}{\theta_i}\right)
  \left\lvert\deriv{g(\epsilon;\theta)}{\epsilon}\right\rvert
  + \p{g(\epsilon;\theta);\theta}
  \left\lvert\deriv{g(\epsilon;\theta)}{\epsilon}\right\rvert
  \textup{Tr}\left[\deriv{g(\epsilon;\theta)}{\epsilon}^{-1}\pderivwsq{g(\epsilon;\theta)}{\theta_i}{\epsilon}\right]
\\
    &=\pderivw{\p{\bfv{x};\theta}}{\theta_i}
    + \pderivw{\p{\bfv{x};\theta}}{\bfv{x}}\pderivw{g(\epsilon;\theta)}{\theta_i}  + \p{\bfv{x};\theta}
  \textup{Tr}\left[\deriv{g(\epsilon;\theta)}{\epsilon}^{-1}\pderivwsq{g(\epsilon;\theta)}{\theta_i}{\epsilon}\right],
\end{aligned}
\end{equation}
where we used the matrix identity
$\pderivw{\lvert X\rvert}{y} = \lvert
X\rvert\textup{Tr}\left[X^{-1}\pderivw{X}{y}\right]$, e.g.\ see the
matrix cookbook \citep{matrixcookbook}. Also, note that we canceled
the $\left\lvert\deriv{g(\epsilon;\theta)}{\epsilon}\right\rvert$
term, because the total sum is $0$, so division by a constant does not
affect the equation.  Further, note that
$\deriv{g(\epsilon;\theta)}{\epsilon}^{-1} =
\deriv{S(\bfv{x};\theta)}{\bfv{x}}$ from the inverse function theorem,
because $g$ and $S$ are inverses. Finally, note that
$\textup{Tr}\left[\deriv{S(\bfv{x};\theta)}{\bfv{x}}\pderivwsq{g(\epsilon;\theta)}{\theta_i}{\epsilon}\right]
=
\textup{Tr}\left[\pderivwsq{g(\epsilon;\theta)}{\theta_i}{\epsilon}\deriv{S(\bfv{x};\theta)}{\bfv{x}}\right]$
because $\textup{Tr}\left[AB\right] = \textup{Tr}\left[BA\right]$ for
arbitrary matrices $A$ and $B$. Combining the results in
Eqs.~(\ref{RPflowcont})~and~(\ref{RPflownablaexp}) we see that they
are the same as Eq.~(\ref{eq:pepszero}); hence, reparameterization
gradients satisfy the transport equation for arbitrary invertible
transformations $g$ and $S$.

\paragraph{Implicit reparameterization gradient:}
The flow field for implicit reparameterization gradients
\citep{figurnov2018implicitRP} is given by
\begin{equation}
\label{implicitflow}
  \bfv{u}_{\theta_i}(\bfv{x}) =
  -\left(\nabla_{\bfv{x}}S(\bfv{x};\theta)\right)^{-1}\nabla_{\theta_i}S(\bfv{x};\theta) =
  -\left(\pderivw{S(\bfv{x};\theta)}{\bfv{x}}\right)^{-1}\pderivw{S(\bfv{x};\theta)}{\theta_i}.
\end{equation}
We will show that the transport equation is satisfied by showing the
equivalence to the flow field for the explicit reparameterization gradient
case. We perform the reverse
of the implicit reparameterization gradient derivation by
\citet{figurnov2018implicitRP}. We can write $\bfv{x} =
g(S(\bfv{x};\theta);\theta)$. Then, by taking the derivative w.r.t.\
$\theta_i$ of both sides, the left-hand side will be $0$, because it
is independent of $\theta_i$, and the right-hand side will give us our
desired equation:
\begin{equation}
\label{reparimplflow}
\begin{aligned}
  \deriv{}{\theta_i}\biggl(g(S(\bfv{x};\theta);\theta)\biggr)
  &=
  \pderiv{g(\epsilon;\theta)}{\theta_i}{\epsilon=S(\bfv{x};\theta)}
  +\pderivw{g(\epsilon;\theta)}{\epsilon}\pderivw{S(\bfv{x};\theta)}{\theta_i}
  = 0\\ \Rightarrow
  \pderiv{g(\epsilon;\theta)}{\theta_i}{\epsilon=S(\bfv{x};\theta)} &=
  -\pderivw{g(\epsilon;\theta)}{\epsilon}\pderivw{S(\bfv{x};\theta)}{\theta_i}.
\end{aligned}
\end{equation}
Now, note that based on the inverse function theorem
$\left(\pderivw{S(\bfv{x};\theta)}{\bfv{x}}\right)^{-1} =
\pderivw{g(\epsilon;\theta)}{\epsilon}$, so
Eqs.~(\ref{implicitflow})~and~(\ref{reparimplflow}) are the same. Note
that
$\pderiv{g(\epsilon;\theta)}{\theta_i}{\epsilon=S(\bfv{x};\theta)}$
was the flow $\bfv{u}_{\theta_i}(\bfv{x})$ for the explicit
reparameterization gradient in the previous section. Therefore, the
implicit reparameterization gradient also explicitly satisfies the
transport equation.

\paragraph{Generalized reparameterization gradient:}
Our work also generalizes the generalized reparameterization gradient
(GRP) \citep{ruiz2016generalizedRP}. Unlike standard RP, in GRP, the
distribution for $\epsilon$ may also depend on $\theta$, i.e.\
$\p{\epsilon;\theta}$.  \citet{ruiz2016generalizedRP} showed that GRP
can be written with the equation:
\begin{equation}
\label{eq:genRP}
\begin{aligned}
  \deriv{}{\theta_i}\expectw{\bfv{x}\sim\p{\bfv{x};\theta}}{\phi(\bfv{x})}
  = 
  \mathbb{E}_{\bfv{x}\sim p(\bfv{x};\theta)}\biggl[
    \bfv{h}_{\theta_i}\bigl(S(x;\theta);\theta\bigr)
    \cdot\nabla_{\bfv{x}}\phi(\bfv{x})&
    \\+ 
    \biggl(
    \nabla_{\bfv{x}}\log\p{\bfv{x};\theta}\cdot\bfv{h}_{\theta_i}\left(S(\bfv{x};\theta);\theta\right)
    +\pderivw{\log\p{\bfv{x};\theta}}{\theta_i}
    + &u_{\theta_i}(S(\bfv{x};\theta);\theta)
    \biggr)\phi(\bfv{x})\biggr] \\ = 
  \mathbb{E}_{\bfv{x}\sim p(\bfv{x};\theta)}\biggl[
    \bfv{h}_{\theta_i}\bigl(S(x;\theta);\theta\bigr)
    \cdot\nabla_{\bfv{x}}\phi(\bfv{x})&
    \\+ 
    \frac{1}{p\bigl(\bfv{x};\theta\bigr)}\biggl(
    \nabla_{\bfv{x}}\p{\bfv{x};\theta}\cdot\bfv{h}_{\theta_i}\left(S(\bfv{x};\theta);\theta\right)
    + \p{\bfv{x};\theta}u_{\theta_i}(S(\bfv{x};&\theta);\theta)
    +\pderivw{\p{\bfv{x};\theta}}{\theta_i}
    \biggr)\phi(\bfv{x})\biggr],
\end{aligned}
\end{equation}
where
$\bfv{h}_{\theta_i}(\epsilon;\theta) =
\pderivw{g(\epsilon;\theta)}{\theta_i}$,
$u_{\theta_i}(\epsilon;\theta) = \pderivw{\log
  J(\epsilon;\theta)}{\theta_i}$, and
$J(\epsilon;\theta) = \left\lvert
  \pderivw{g(\epsilon;\theta)}{\epsilon}\right\rvert$. Note that
$u_{\theta_i}(\epsilon;\theta)~\neq~\bfv{u}_{\theta_i}(\bfv{x})$ as in
our previous notation, and the cause of the confusing notation is that we
chose to use the same notation in Eq.~(\ref{eq:genRP}) as the work by
\citet{ruiz2016generalizedRP}.  Comparing the $\nabla_{\bfv{x}}\phi(\bfv{x})$
terms in
Eqs.~(\ref{eq:genRP})~and~(\ref{flowest}) it is clear that we must
have
\begin{equation}
  \bfv{u}_{\theta_i}(\bfv{x})~=~\bfv{h}_{\theta_i}(S(\bfv{x};\theta);\theta)~=~\pderiv{g(\epsilon;\theta)}{\theta_i}{\epsilon=S(\bfv{x};\theta)}.
\end{equation}
By expanding the divergence term in Eq.~(\ref{flowest}),
$\nabla_{\bfv{x}}\cdot~\bigl(\p{\bfv{x};\theta}\bfv{u}_{\theta_i}(\bfv{x})\bigr)~=~\nabla_{\bfv{x}}\p{\bfv{x};\theta}\cdot\bfv{u}_{\theta_i}(\bfv{x})~+~\p{\bfv{x};\theta}\nabla_{\bfv{x}}\cdot\bfv{u}_{\theta_i}(\bfv{x})$,
and comparing the $\phi(\bfv{x})$ terms, one sees that to achieve equivalence between the two equations, we must have
\begin{equation}
  u_{\theta_i}(S(\bfv{x};\theta);\theta) =
  \nabla_{\bfv{x}}\cdot\bfv{u}_{\theta_i}(\bfv{x}) = \nabla_{\bfv{x}}\cdot
  \bfv{h}_{\theta_i}(S(\bfv{x};\theta);\theta).
\end{equation}
 The left-hand side
term is
\begin{equation}
\begin{aligned}
  u_{\theta_i}(S(\bfv{x};\theta);\theta) =
  \pderiv{\log
    J(\epsilon;\theta)}{\theta_i}{\epsilon=S(\bfv{x};\theta)}
  &=
  \left.\pderivw{}{\theta_i}\left(\log
    \left\lvert
  \pderivw{g(\epsilon;\theta)}{\epsilon}\right\rvert\right)
\right\rvert_{\epsilon=S(\bfv{x};\theta)}\\
&=
\textup{Tr}\biggl[
\pderivw{g(\epsilon;\theta)}{\epsilon}^{-1}
\pderivsq{g(\epsilon;\theta)}{\theta_i}{\epsilon}{\epsilon=S(\bfv{x};\theta)}
\biggr]\\
&=
\textup{Tr}\biggl[
\pderivsq{g(\epsilon;\theta)}{\theta_i}{\epsilon}{\epsilon=S(\bfv{x};\theta)}
\pderivw{S(\bfv{x};\theta)}{\bfv{x}}
\biggr]
,
\end{aligned}
\end{equation}
where we used the fact that
$\pderivw{g(\epsilon;\theta)}{\epsilon}^{-1} = \pderivw{S(\bfv{x};\theta)}{\bfv{x}}$
from the inverse function theorem, as well as the matrix identities
$\pderivw{}{y}\log|X| = \textup{Tr}[X^{-1}\pderivw{X}{y}]$ and
$\textup{Tr}[AB] = \textup{Tr}[BA]$, e.g.\ see the matrix cookbook
\citep{matrixcookbook}. We will show that the right-hand side term is
the same:
\begin{equation}
\nabla_{\bfv{x}}\cdot
\bfv{h}_{\theta_i}(S(\bfv{x};\theta);\theta)
=
\nabla_{\bfv{x}}\cdot
\pderiv{g(\epsilon;\theta)}{\theta_i}{\epsilon=S(\bfv{x};\theta)}
=
\textup{Tr}\biggl[
\pderivsq{g(\epsilon;\theta)}{\theta_i}{\epsilon}{\epsilon=S(\bfv{x};\theta)}
\pderivw{S(\bfv{x};\theta)}{\bfv{x}}
\biggr],
\end{equation}
which is the desired result. Therefore, the GRP gradient is a special
case of the flow gradient estimator in Eq.~(\ref{flowest}). Note that
in the standard reparameterization gradient derivation, we used the
fact that $\deriv{\p{\epsilon}}{\theta_i} = 0$ to show that the
transport equation holds, and hence that the multiplier for
$\phi(\bfv{x})$ disappears, but in the GRP case, the distribution for
$\epsilon$ depends on $\theta$, so
$\deriv{\p{\epsilon;\theta}}{\theta_i} \neq 0$, and the $\phi(\bfv{x})$
term does not disappear. Finally, note that it is an open question whether the
reverse may also be true---could it be that the GRP and flow gradient
estimator spaces are equal? To show that they are equal,
one would have to find a generalized reparameterization corresponding
to each arbitrary $\bfv{u}_{\theta_i}(\bfv{x})$. However,
we believe that if at all possible,
the reparameterization corresponding to some complicated flow field
may be quite bizarre, while in the flow framework, one just has to do a
dot product between the flow and the gradient to compute the estimator.

\subsection{Flow gradients with discontinuities}
\label{flowdiscontapp}

In Eq.~(\ref{appdivF}) when applying the divergence theorem, we
assumed that
$\bfv{F} = \p{\bfv{x};\theta}\bfv{u}_{\theta_i}(\bfv{x})\phi(\bfv{x})$
is a continuous piecewise differentiable vector field, so that the
divergence theorem would hold. Moreover, in
Eq.~(\ref{plusmintodtheta}), we assumed that
$\deriv{}{\theta}\int~\p{\bfv{x};\theta}\phi(\bfv{x})\textup{d}\bfv{x}~=~\int~\deriv{\p{\bfv{x};\theta}}{\theta}\phi(\bfv{x})\textup{d}\bfv{x}$,
which will not be true when $\p{\bfv{x};\theta}$ has discontinuities.
Here, we extend the theory to all possible discontinuities. Our work
casts previous work on RP gradients for discontinuous functions
$\phi(\bfv{x})$ into the flow gradient framework
\citep{lee2018rpnondiff,cong2019go}, but also characterizes all other
possible discontinuities, which have not been considered in the
literature yet. We consider first what to do about discontinuities in
$\bfv{F}$, then discuss discontinuities in $\p{\bfv{x};\theta}$ (which
would automatically mean that $\bfv{F}$ is also discontinuous).

\paragraph{Discontinuities in $\bfv{F} = \p{\bfv{x};\theta}\bfv{u}_{\theta_i}(\bfv{x})\phi(\bfv{x})$:}

A well-known result states that if there are discontinuities in
$\bfv{F}$ appearing on the surface $M$, then the divergence theorem
has to be modified by adding the jump across the surface, giving the
equation
\begin{equation}
\label{discontdivergence}
\int_V \nabla_{\bfv{x}}\cdot\bfv{F}\textup{d}V = \int_S \bfv{F}\cdot\textup{d}\bfv{S}
+ \int_M \Delta \bfv{F}\cdot\textup{d}\bfv{M},
\end{equation}
where $S$ is the surface enclosing the volume $V$, $M$ is the surface
inside $V$ where the discontinuities occur, and $\Delta\bfv{F}$ is the
difference in the vector field $\bfv{F}$ between the two sides of the
discontinuity. We give a short proof of this claim:

Partition the volume $V$ into $K$ disjoint regions, s.t.\ $\bfv{F}$ is
continuous in each region $V_i$, and the discontinuity surface $M$ is
contained at the surface boundaries of the disjoint regions. In this
case, because $\bfv{F}$ is continuous in each region $V_i$, we can
apply the divergence theorem for each region separately, and sum
the results. 
\begin{equation}
\label{discontvolumes}
  \sum_{i=1}^K\int_{V_i} \nabla_{\bfv{x}}\cdot\bfv{F}\textup{d}V =
  \sum_{i=1}^K\int_{S_i} \bfv{F}\cdot\textup{d}\bfv{S}.
\end{equation}
It will turn out, that the flow across the inner surfaces will cancel,
unless there is a discontinuity, which gives the additional term in
Eq.~(\ref{discontdivergence}).  We write the surface integral, as the
sum over the outer surfaces $S_{out,i}$ (i.e.\ exterior to $V$), which
add up to $S$, and the inner surfaces $S_{in,i}$ (i.e.\ interior to
$V$), which enclose $M$ as well as the other region boundaries.  Then,
the right-hand side term becomes:
\begin{equation}
\label{volumesintomanifold}
\begin{aligned}
  \sum_{i=1}^K\int_{S_i} \bfv{F}\cdot\textup{d}\bfv{S} &=
  \sum_{i=1}^K\int_{S_{out,i}} \bfv{F}\cdot\textup{d}\bfv{S} +
  \sum_{i=1}^K\int_{S_{in,i}} \bfv{F}\cdot\textup{d}\bfv{S}\\
  &= \int_S \bfv{F}\cdot\textup{d}\bfv{S}
+ \int_M \Delta \bfv{F}\cdot\textup{d}\bfv{M},
\end{aligned}
\end{equation}
where each outer surface only appears for one of the volumes $V_i$,
and gives the first term, while each inner surface belongs to two
regions $V_j$ and $V_k$, which gives rise to the difference in the
vector fields because the surface normal pointing outward from the
volume is in opposite directions for the two separate volumes. Note
that if there is no discontinuity at an inner surface, then the
difference in vector fields is $0$, and it can be ignored, so only the
integral across $M$ matters.  Finally, note that
$\sum_{i=1}^K\int_{V_i} \nabla_{\bfv{x}}\cdot\bfv{F}\textup{d}V =
\int_{V} \nabla_{\bfv{x}}\cdot\bfv{F}\textup{d}V,$ which concludes the
proof.  To estimate the surface integral, one could sample additional
points on the surface $M$, and apply importance sampling to estimate
the integral, e.g.\ as done by \citet{lee2018rpnondiff}. Once, we have estimated
the surface integral across $M$, the flow gradient estimator in
Eq.~(\ref{flowest}), which we denote here by $E$, has to be modified
to make it unbiased. In particular, in Eq.~(\ref{flowgradeq}) we assumed
that the divergence will be $0$, but in the discontinuous case, it will
instead equal $G = \int_M \Delta \bfv{F}\cdot\textup{d}\bfv{M}$, so
we must subtract it from the estimator, to make it unbiased, i.e.\
we use the estimator $E-G$, or in practice, not exactly $G$, but an
estimator for $G$:
\begin{equation}
\label{discontflowest}
\begin{aligned}
  &\deriv{}{\theta_i}\expectw{\bfv{x}\sim\p{\bfv{x};\theta}}{\phi(\bfv{x})}\\
  &=
  \expectw{\bfv{x}\sim q(\bfv{x})}{
    \frac{\p{\bfv{x};\theta}}{q\left(\bfv{x}\right)}
    \bfv{u}_{\theta_i}(\bfv{x})\cdot\nabla_{\bfv{x}}\phi(\bfv{x})
    +
    \frac{1}{q\left(\bfv{x}\right)}\biggl(\nabla_{\bfv{x}}\cdot
  \bigl(\p{\bfv{x};\theta}\bfv{u}_{\theta_i}(\bfv{x})\bigr)
  + \deriv{\p{\bfv{x};\theta}}{\theta_i}\biggr)\phi(\bfv{x})
}\\
&-
\expectw{\bfv{x}\sim q_M(\bfv{x})}
{\frac{1}{q_M(\bfv{x})}
  \Delta\biggl(\p{\bfv{x};\theta}\phi(\bfv{x})\bfv{u}_{\theta_i}(\bfv{x})\biggr)
  \cdot\hat{\bfv{m}}(\bfv{x})},
\end{aligned}
\end{equation}
where $q_M$ is a probability distribution on the surface $M$, where
the discontinuity occurs, $\Delta(\cdot)$ computes the change across
the discontinuity, and $\hat{\bfv{m}}(\bfv{x})$ is the surface
normal vector on $M$ pointing in the opposite direction in which $\Delta$
is computed.

Another intuitive way to prove the claim in
Eq.~(\ref{discontdivergence}) would be to define a parameterized
smooth relaxation of the vector field $\bfv{F}(\bfv{x})$, where the
discontinuous steps are swapped with smooth transitions, given by
$\tilde{\bfv{F}}(\bfv{x};\gamma)$, where
$\lim_{\gamma\to\infty}\tilde{\bfv{F}}(\bfv{x};\gamma) =
\bfv{F}(\bfv{x})$. In this case, the divergence theorem can be applied
on $\tilde{\bfv{F}}$, and taking the limit will give the theorem for
$\bfv{F}$. It would turn out that the integral of
$\nabla_{\bfv{x}}\cdot\tilde{\bfv{F}}$ over the discontinuous regions
$M$ would not disappear as the tightness of the smooth transitions is
increased, because while the region of integration shrinks, the
magnitude of the gradient would also increase, and the integral across
the discontinuity will tend to the change in $\bfv{F}$.

We can now cast previous works \citep{lee2018rpnondiff,cong2019go}
about discontinuities into the flow gradient framework, and show that
they arise by considering a discontinuity in $\bfv{F}$. In particular,
they consider the special case of a discontinuity in $\phi(\bfv{x})$,
which we will explain next.

\paragraph{Discontinuities in $\phi(\bfv{x})$:}
When $\phi(\bfv{x})$, has discontinuities with jumps
$\Delta\phi(\bfv{x})$ at the surface $M$, then $\Delta\bfv{F}$ in
Eq.~(\ref{discontdivergence}) is
$\Delta(\p{\bfv{x};\theta}\bfv{u}_{\theta_i}(\bfv{x})\phi(\bfv{x})) =
\p{\bfv{x};\theta}\bfv{u}_{\theta_i}(\bfv{x})\Delta\phi(\bfv{x})$.  We
will explain that this gives rise to the reparameterization gradients
with discontinuous models by \citet{lee2018rpnondiff}. In their
Theorem 1, they gave the equation
\begin{equation}
\label{leediscont}
  \nabla_{\theta_i}\textup{ELBO} =
  \expectw{\epsilon\sim\p{\epsilon}}{\nabla_{\theta}h(\epsilon;\theta)}
  + \sum_{i=1}^K\int_{S_i}\p{\epsilon}h(\epsilon;\theta)
  \bfv{V}(\epsilon;\theta)\cdot\textup{d}\bfv{S}_{\epsilon},
\end{equation}  
where
$\textup{ELBO} =
\expectw{\epsilon\sim\p{\epsilon}}{h(\epsilon;\theta)}$, the $S_i$ are
the surfaces of $K$ disjoint continuous volumes (similar to our proof
of the discontinuous divergence theorem in Eq.~(\ref{discontvolumes})
), $h(\epsilon;\theta) = \phi(g(\epsilon;\theta))$, and
$V(\epsilon;\theta) =
\pderiv{S(\bfv{x};\theta)}{\theta_i}{\bfv{x}=g(\epsilon;\theta)}$. To
compare their equation to our derivation, we must change the
coordinates from the $\epsilon$ space to the $\bfv{x}$ space.  The
main difference is the change in the direction of the surface normal
$\hat{\bfv{n}}_\epsilon$ on each $S_i$ to $\hat{\bfv{n}}_{\bfv{x}}$,
as well as the change in the vector field
$\bfv{V}(\epsilon;\theta)$. In our derivation, we will first transform
the surface integral into a volume integral by using the divergence
theorem, and then perform the change in coordinates, as this allows to
ignore the change in orientation of the surface normal. The surface
integral is equal to the volume integral of the divergence:
\begin{equation}
\int_{S_i}\p{\epsilon}h(\epsilon;\theta)
\bfv{V}(\epsilon;\theta)\cdot\textup{d}\bfv{S}_{\epsilon}
=
\int_{V_i}
\nabla\cdot\biggl(\p{\epsilon}h(\epsilon;\theta)
\bfv{V}(\epsilon;\theta)\biggr)\textup{d}V_\epsilon.
\end{equation}
The vector field $\bfv{V}(\epsilon;\theta)$ denotes a rate of
change in the $\epsilon$ space; to change the coordinates to
the $\bfv{x}$ space, we apply the chain rule: \[\bfv{u}(\bfv{x};\theta)
= \pderivw{g(\epsilon;\theta)}{\epsilon}\bfv{V}(\epsilon;\theta) =
\pderivw{g(\epsilon;\theta)}{\epsilon}\pderiv{S(\bfv{x};\theta)}{\theta_i}{\bfv{x}=g(\epsilon;\theta)} = -\pderiv{g(\epsilon;\theta)}{\theta_i}{\epsilon=S(\bfv{x};\theta)},\]
from Eq.~(\ref{reparimplflow}).

When performing a change of coordinates for the divergence, we can
apply the Voss-Weyl formula \citep{grinfeld2013introduction}
\begin{equation}
  \nabla_\epsilon\cdot\bfv{V}(\epsilon;\theta) =
  \frac{1}{|J|}\nabla_{\bfv{x}}\cdot\left(|J|\bfv{u}(\bfv{x};\theta)\right),
\end{equation}
where $|J|$ is the determinant of the Jacobian $J =
\pderivw{S(\bfv{x})}{\bfv{x}}$. Similarly, the change in
coordinates for the volume element is given by $\textup{d}V_\epsilon =
|J|\textup{d}V_{\bfv{x}}$. Combining the results, we have
\begin{equation}
\begin{aligned}
\int_{V_i}
\nabla_{\epsilon}\cdot\biggl(\p{\epsilon}h(\epsilon;\theta)
\bfv{V}(\epsilon;\theta)\biggr)\textup{d}V_\epsilon
&=
-\int_{V_i}
\nabla_{\bfv{x}}\cdot\biggl(\left\lvert\pderivw{S(\bfv{x})}{\bfv{x}}\right\rvert
\p{\epsilon}\phi(\bfv{x})
\pderiv{g(\epsilon;\theta)}{\theta_i}{\epsilon=S(\bfv{x};\theta)}\biggr)\textup{d}V_{\bfv{x}}\\
&=
-\int_{V_i}
\nabla_{\bfv{x}}\cdot\biggl(\p{\bfv{x};\theta}\phi(\bfv{x})
\pderiv{g(\epsilon;\theta)}{\theta_i}{\epsilon=S(\bfv{x};\theta)}\biggr)\textup{d}V_{\bfv{x}},
\end{aligned}
\end{equation}
where $\phi(\bfv{x}) = h(S(\bfv{x}))$, as
$h(\epsilon;\theta) := \phi(g(\epsilon;\theta))$, and
$\left\lvert\pderivw{S(\bfv{x})}{\bfv{x}}\right\rvert \p{\epsilon} =
\p{\bfv{x};\theta}$, and the minus sign comes because
$\bfv{u}(\bfv{x};\theta) =
-\pderiv{g(\epsilon;\theta)}{\theta_i}{\epsilon=S(\bfv{x};\theta)}$. Finally,
we apply the divergence theorem again:
\begin{equation}
-\int_{V_i}
\nabla_{\bfv{x}}\cdot\biggl(\p{\bfv{x};\theta}\phi(\bfv{x})
\pderiv{g(\epsilon;\theta)}{\theta_i}{\epsilon=S(\bfv{x};\theta)}\biggr)\textup{d}V_{\bfv{x}}
=
-\int_{S_i}
\p{\bfv{x};\theta}\phi(\bfv{x})
\pderiv{g(\epsilon;\theta)}{\theta_i}{\epsilon=S(\bfv{x};\theta)}\cdot\textup{d}\bfv{S}_{\bfv{x}}.
\end{equation}
Note that $\pderiv{g(\epsilon;\theta)}{\theta_i}{\epsilon=S(\bfv{x};\theta)}
= \bfv{u}_{\theta_i}(\bfv{x})$ is the flow for the reparameterization gradient
in Eq.~(\ref{RPflowcont}). Therefore, we have 
\begin{equation}
-\sum_{i=1}^K\int_{S_i}
\p{\bfv{x};\theta}\phi(\bfv{x})
\pderiv{g(\epsilon;\theta)}{\theta_i}{\epsilon=S(\bfv{x};\theta)}\cdot\textup{d}\bfv{S}_{\bfv{x}} = -\int_M \p{\bfv{x};\theta}\Delta\phi(\bfv{x}) \bfv{u}_{\theta_i}(\bfv{x})\cdot\textup{d}\bfv{M},
\end{equation}
where the change to $\Delta\phi(\bfv{x})$ instead of summing the flow through
each surface is analogous to 
what we explained in Eq.~(\ref{volumesintomanifold}): the integral
across the inner surfaces occurs in two of the disjoint volumes, which
gives rise to integrating the change in the vector field. We can now
see that their estimator, given in Eq.~(\ref{leediscont}), is analogous
to our estimator for the discontinuous case in Eq.~(\ref{discontflowest}),
but performed in the $\epsilon$ space, and while assuming a perfect
reparameterization.

\paragraph{GO gradient estimator:} Next, we show that the GO gradient
estimator for discrete variables \citep{cong2019go} can also be derived
from our framework. We illustrate the equivalence on a simplified
1-dimensional case, which is straightforward to generalize to
arbitrary dimensions.  \citet{cong2019go} consider a discrete variable
$y\in\{0,...,\infty\}$, a function $\phi(y)$, and a probability
distribution $p(y;\theta)$.  Moreover, they define
$Q(y;\theta) = \sum_{i=0}^yp(i;\theta)$. Then, they provide the
gradient estimator:
\begin{equation}
  \deriv{}{\theta}\expectw{y\sim\p{y;\theta}}{\phi(y)} =
  \expectw{y\sim\p{y;\theta}}{\frac{-1}{\p{y;\theta}}\deriv{Q(y;\theta)}{\theta}
    \biggl(\phi(y+1)-\phi(y)\biggr)},
\end{equation}
We will show that this estimator can be derived from the flow theory.
Consider a continuous space of $x$, with $\phi(x) = \phi(y)$, when
$y\leq x < y+1$, and
$\int_{y}^{y+1}\p{x;\theta}\textup{d}x = \p{y;\theta}$, which occurs
when $\p{x;\theta} = \p{y;\theta}$ if $y\leq x < y+1$, then the
expectation over $x$ and $y$ give the same results. Next, we define a
flow field $\bfv{u}_{\theta_i}(x)$ that satisfies the transport
equation, i.e.\ the $\phi(x)$ term in the flow gradient estimator will
disappear, and only the $\nabla_{\bfv{x}}\phi(x)$ terms will
remain. But $\nabla_{\bfv{x}}\phi(x) = 0$ everywhere, so that term
also disappears. It will only be necessary to estimate the flow
$\bfv{u}_{\theta_i}(x)$ through the discontinuous boundaries at each
$x=y+1$, i.e.\ the $q_M$ term in Eq.~(\ref{discontflowest}). We apply
the divergence theorem on
$\nabla_{\bfv{x}}\cdot\left(\p{x;\theta}\bfv{u}_{\theta_i}(x)\right)$:
\begin{equation}
\begin{aligned}
  \int_0^{y+1}\nabla_{\bfv{x}}\cdot\left(\p{x;\theta}\bfv{u}_{\theta_i}(x)\right)
  \textup{d}x =
  \int_S \p{x;\theta}\bfv{u}_{\theta_i}(x)\cdot\textup{d}\bfv{S}\\
  = \p{x=y+1;\theta}\bfv{u}_{\theta_i}(x=y+1) = \p{y;\theta}\bfv{u}_{\theta_i}(y+1),
\end{aligned}
\end{equation}
where the equation comes from the fact that the only exterior surface
with non-zero $\bfv{u}_{\theta_i}(x)$ is at $x=y+1$, and by definition
$\p{x=y+1;\theta} = \p{y;\theta}$. Moreover, note that because of the
transport equation
$\nabla_{\bfv{x}}\cdot\left(\p{x;\theta}\bfv{u}_{\theta_i}(x)\right) =
-\deriv{\p{x;\theta}}{\theta}$, and hence, we can obtain another
expression for $\p{y;\theta}\bfv{u}_{\theta_i}(y+1)$ by computing the
integral
\begin{equation}
  \int_0^{y+1}\nabla_{\bfv{x}}\cdot\left(\p{x;\theta}\bfv{u}_{\theta_i}(x)\right)
  \textup{d}x
  =
  \sum_{i=0}^{y}-\deriv{\p{i;\theta}}{\theta} = -\deriv{Q(y;\theta)}{\theta}.
\end{equation}
Therefore, we have
$\p{y;\theta}\bfv{u}_{\theta_i}(y+1) =
-\deriv{Q(y;\theta)}{\theta}$. Now, noting that the jump across the
surface at $x=y+1$ is given by $\phi(y+1)-\phi(y)$, and taking into
account of the opposite direction of $\hat{\bfv{m}}$ we have
that
\[\Delta\bfv{F}(x=y+1)\cdot\hat{\bfv{m}}(x=y+1) =
  \deriv{Q(y;\theta)}{\theta}\biggl(\phi(y+1)-\phi(y)\biggr).\]
Plugging into Eq.~(\ref{discontflowest}), while noting that
$q_M(x=y+1) := \p{y;\theta}$, we obtain the GO gradient estimator. In
the original derivation of \citet{cong2019go}, they used an algebraic
derivation based on integration by parts.  The difficulty was how to
extend the derivation for discrete variables.  They extended
it by using an Abel transformation
\citep{abel1895untersuchungen} and summation by parts. Our derivation,
on the other hand, casts this gradient estimator into the flow
gradient framework, and provides a physical insight into the principle
behind the estimator.

\remove{considering that the jump across the
discontinuous manifold $\int_M \Delta \bfv{F}\cdot\textup{d}\bfv{M}$,
where the jump $\Delta\bfv{F} = \Delta(\p{x+1;\theta}\bfv{u}_{\theta_i}(x+1)\phi(x+1)) =
\p{x+1;\theta}\bfv{u}_{\theta_i}(x+1)\Delta\phi(x+1) =
\p{x+1;\theta}\bfv{u}_{\theta_i}(x+1)\left(\phi(x+1) - \phi(x)\right)
= -\deriv{Q(y;\theta)}{\theta}\left(\phi(y+1) - \phi(y)\right)$,
has to be subtracted to obtain an unbiased gradient estimator (which
gets rid of the minus sign on $\deriv{Q(y;\theta)}{\theta}$)}

\paragraph{Discontinuities in $\p{\bfv{x};\theta}$:} When there are
discontinuities in $\p{\bfv{x};\theta}$, then $\bfv{F}$ will be
discontinuous, which is straightforward to deal with based on our
previous example; however, that is not enough, because
$\int
\deriv{\p{\bfv{x};\theta}}{\theta}\phi(\bfv{x})\textup{d}\bfv{x}~\neq~\deriv{}{\theta}\expectw{\bfv{x}\sim\p{\bfv{x};\theta}}{\phi(\bfv{x})}$
in general. To deal with the mismatch, it becomes necessary to split
the domain of integration into piecewise continuous domains, then
perform the integration by taking into account the movement of the
discontinuous boundary. In general, when differentiating an integral
of a function $f(\bfv{x};\theta)$ in a moving domain $D(\theta)$, the
integral is given by
\begin{equation}
  \deriv{}{\theta}\int_{D(\theta)}f(\bfv{x};\theta)\textup{d}\bfv{x} =
  \int_{D(\theta)}\deriv{f(\bfv{x};\theta)}{\theta}\textup{d}\bfv{x}
  + \int_{S}f(\bfv{x};\theta)\bfv{v}(\bfv{x})\cdot\textup{d}\bfv{S},
\end{equation}
where $S$ is the surface around the domain and $\bfv{v}(\bfv{x})$ is the
velocity of the location of the point on the domain, i.e.\
$\deriv{\bfv{x}}{\theta}$, where $\bfv{x}\in S$. Performing the usual split
of the domain of integration into piecewise continuous domains,
and applying the rule for integration under moving boundaries gives
the equation
\begin{equation}
  \deriv{}{\theta}\int \p{\bfv{x};\theta}\phi(\bfv{x})\textup{d}\bfv{x} =
  \int \deriv{\p{\bfv{x};\theta}}{\theta}\phi(\bfv{x})\textup{d}\bfv{x}
  + \int_{C}\phi(\bfv{x})\Delta\p{\bfv{x};\theta}\bfv{v}(\bfv{x})\cdot\textup{d}\bfv{S},
\end{equation}
where $C$ is the surface where the discontinuities occur, and
$\Delta\p{\bfv{x};\theta}$ is the change in the probability density
across the surfaces, computed in the opposite direction to the surface normal
$\textup{d}\bfv{S}$. Adding the correction factor based on an estimate
of the integral across $C$ will unbias the gradient estimator. The
same trick was applied for
deriving the estimator for RP gradients under discontinuities
\citep{lee2018rpnondiff}.

\section{Slice integral importance sampling}
\label{sliceintegral}

\begin{figure*}
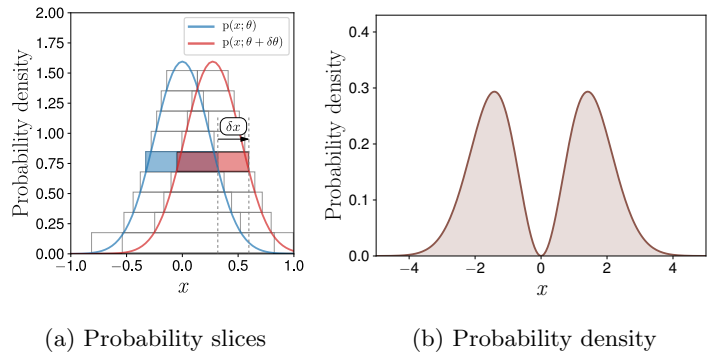

        \centering
	\begin{subfigure}{.244\textwidth}
          \mbox{}\\[0.65em] 
		\includegraphics[width=\textwidth]{LebBars3.pdf}
		\caption{Probability slices}
          \label{lebbars}
        \end{subfigure}
	\begin{subfigure}{.33\textwidth}
          \includegraphics[width=\textwidth]{lebplot.pdf}\\[-0.6em]
          \caption{Probability density}
          \label{ldist}
	\end{subfigure}
	\caption{Slice integral sampling motivation and distribution for
          a Gaussian base distribution.}
          \label{lebbarplot}
\end{figure*}

From Theorem~\ref{lrunique} we saw that unlike the RP gradient case,
the weighting $\psi$ for function values $\phi(\bfv{x})$ with
$\bfv{x} \sim \p{\bfv{x};\theta}$ to obtain an unbiased estimator for
the gradient $\deriv{}{\theta}\expect{\phi(\bfv{x})}$ is unique. The
only option to reduce the variance by changing the weighting would
then be to sample from a different distribution
$q(\bfv{x})$ via importance sampling. Motivated by the
resemblance of the ``boxes" theory in Sec.~\ref{boxtheor} to the
Riemann integral, we propose to sample horizontal slices of
probability mass resembling the Lebesgue integral. Such an approach
appears attractive, because if the location of the slice is moved by
modifying the parameters of the distribution (e.g., by changing the
mean), then the derivative of the expected value of the integral over
the slice will depend only on the value at the edges of the slice
(because the probability density in the middle would not change). To
obtain the gradient estimator, it will only be necessary to compute
the probability density $q_{L}(\bfv{x};\theta)$. We derive such a
``slice integral" distribution corresponding to the Gaussian
distribution. The method resembles the seminal work by
\citet{neal2003slice} on slice sampling in Markov chain Monte Carlo methods. We call our
new distribution the L-distribution, and it is plotted in
Fig.~\ref{ldist}. In retrospect, it turned out that this distribution
is optimal under the assumption that $\phi(x)$ is linear.

\paragraph{Derivation of the pdf of the L-distribution:} One way to
sample whole slices of a probability distribution would be to sample a
height $h$ between 0 and $p_{\mathrm{max}}$ proportionally to the probability
mass at that height. The probability mass at a height $h$ is just
given by $2|x-\mu|$ where $x$ is such that $\p{x;\mu,\sigma} = h$,
i.e., $2|x-\mu|$ is the distance between the edges of
$\p{x;\mu,\sigma}$. The probability mass corresponding to $x$ is then
given by $2|x-\mu|\textup{d}h$. Performing a change of coordinates to the
$x$-domain, and splitting the mass between the two edges of the slice, we get
$|x-\mu|\textup{d}h = |x-\mu|\left|\deriv{\p{x;\mu,\sigma}}{x}\right|\textup{d}x$.  This
gives a closed-form normalized pdf for the L-distribution:

\begin{equation}
  \begin{aligned}
  q_{L}(x;\mu,\sigma) = |x-\mu|\left|\deriv{\p{x;\mu,\sigma}}{x}\right|
  &= |x-\mu|\p{x;\mu,\sigma}\frac{|x-\mu|}{\sigma^2}\\
  &=\frac{|x-\mu|^2}{\sqrt{2\pi}~\sigma^3}
  \exp\left(\frac{-(x-\mu)^2}{2\sigma^2}\right).
\end{aligned}
\label{maxwell}
\end{equation}

One can recognize that Eq.~(\ref{maxwell}) is actually just a
Maxwell-Boltzmann distribution reflected about the origin with the
probability mass split between the two sides.

\paragraph{Sampling from the L-distribution:} To sample from this
distribution, it is necessary to sample points proportionally to the
length of the slices. It suffices to sample uniformly in the area
under the curve in the space augmented with the height dimension $h$,
then selecting the slice on which the sampled point lies. This can be
achieved with the three steps: 1) sample a point from the base
distribution: $x_s \sim \p{x;\mu,\sigma}$, 2) sample a height:
$h \sim \textup{unif}\left(0,\p{x_s;\mu,\sigma}\right)$, 3) sample $x$ from one
of the two
edges of the slice $p^{-1}(h;\mu,\sigma) = \{x:\p{x;\theta} = h\}$, where
$p^{-1}(h)$ inverts the pdf, and computes the locations of $x$
that give a probability density $h$. For the L-distribution, this can be
achieved by sampling $\epsilon_x \sim \mathcal{N}(0,1)$ and
$\epsilon_h \sim \textup{unif}(0,1)$ and transforming these by the equation:
\begin{equation}
  x = \mu \pm\sigma\sqrt{-2\log(\epsilon_h) + \epsilon_x^2}~.
\end{equation}
Now it is straightforward to obtain the LR gradient estimator:
\begin{equation}
  \begin{aligned}
  \deriv{}{\mu}\expectw{x\sim\p{x;\theta}}{\phi(x)} =
  \expectw{x\sim q(x;\theta)}{\frac{\deriv{p}{\mu}}{q(x;\theta)}\phi(x)} &=
  \expectw{x\sim q(x;\theta)}{\frac{1}{x-\mu}\phi(x)} \\
  &= \expect{\frac{\textup{sgn}(x-\mu)}{\sigma\sqrt{-2\log(\epsilon_h) + \epsilon_x^2}}\phi(x)}.
  \end{aligned}
\end{equation}

\FloatBarrier
\newpage
\bibliography{unifgrad}
\bibliographystyle{apalike}


\end{document}